\documentclass[journal]{IEEEtran}

\usepackage{ifpdf}
\usepackage{cite}
\ifCLASSINFOpdf
  \usepackage[pdftex]{graphicx}
  \graphicspath{{./images/}}
  \DeclareGraphicsExtensions{.pdf,.jpeg,.png,.jpg}
\else
\fi
\usepackage{amsmath, amssymb}
\usepackage{algorithm}
\usepackage{algpseudocode}
\usepackage{array}
\ifCLASSOPTIONcompsoc
 \usepackage[caption=false,font=normalsize,labelfont=sf,textfont=sf]{subfig}
\else
 \usepackage[caption=false,font=footnotesize]{subfig}
\fi

\usepackage{listings}
\usepackage[colorlinks=true, allcolors=blue]{hyperref}
\usepackage{pdflscape}

\usepackage{stfloats}
\usepackage{url}
\begin{document}
\title{A Numerical Gradient Inversion Attack \\ in Variational Quantum Neural-Networks}
%
%
%

\author{Georgios~Papadopoulos$^1$,
        Shaltiel~Eloul$^1$,
        Yash~Satsangi$^1$,
        Jamie~Heredge$^2$,
        Niraj~Kumar$^2$,
        Chun-Fu~Chen$^2$,
        and~Marco~Pistoia$^2$
\thanks{$^1$Global Technology Applied Research, JPMorgan Chase, London, E14 5JP, UK}
\thanks{$^2$Global Technology Applied Research, JPMorgan Chase, New York, NY 10001, USA}}

\maketitle

\begin{sloppypar}

\begin{abstract} 
 The loss landscape of Variational Quantum Neural Networks (VQNNs) is characterized by local minima that grow exponentially with increasing qubits. Because of this, it is more challenging to recover information from model gradients during training compared to classical Neural Networks (NNs). In this paper we present a numerical scheme that successfully reconstructs input training, real-world, practical data from trainable VQNNs' gradients. Our scheme is based on gradient inversion that works by combining gradients estimation with the finite difference method and adaptive low-pass filtering. The scheme is further optimized with Kalman filter to obtain efficient convergence. Our experiments show that our algorithm can invert even batch-trained data, given the VQNN model is sufficiently \textit{`over-parameterized'}. 
\end{abstract}


\ifCLASSOPTIONpeerreview
\begin{center} \bfseries EDICS Category: 3-BBND \end{center}
\fi
%
\IEEEpeerreviewmaketitle

\section{Introduction}
\IEEEPARstart{Q}{uantum} Computing (QC) experiences significant technological advancements every year\cite{Bravyi2022, Lau2022, 10466774, Pfaendler2024, 2025MS, Liu2025}. This progress is anticipated to influence a wide-range of scientific and industrial applications. Quantum Machine Learning (QML) is one of the domains considered to be early beneficiaries of QC, with development of algorithms such as Quantum Support Vector Machines~\cite{PhysRevLett.113.130503} (QSVM), Quantum Deep Neural Networks~\cite{Beer2020} (QDNN), and Quantum Reinforcement Learning~\cite{Saggio2021} (QRL). QML models are designed to fit patterns in data, similar to classical NNs. QMLs are generally grouped into $a)$ models with fixed quantum circuits \cite{Barnes2019,arxiv.2403.02871,arxiv.2101.11020} such as quantum kernel methods, that do not have variational parameters in the quantum circuit and $b)$ VQNN models, a flexible technique that incorporates trainable parameters within the quantum circuits (ansatz) \cite{Du2022,Cerezo2021}. 

\begin{figure}[!t]
    \centering
        \includegraphics[width=0.48\textwidth]{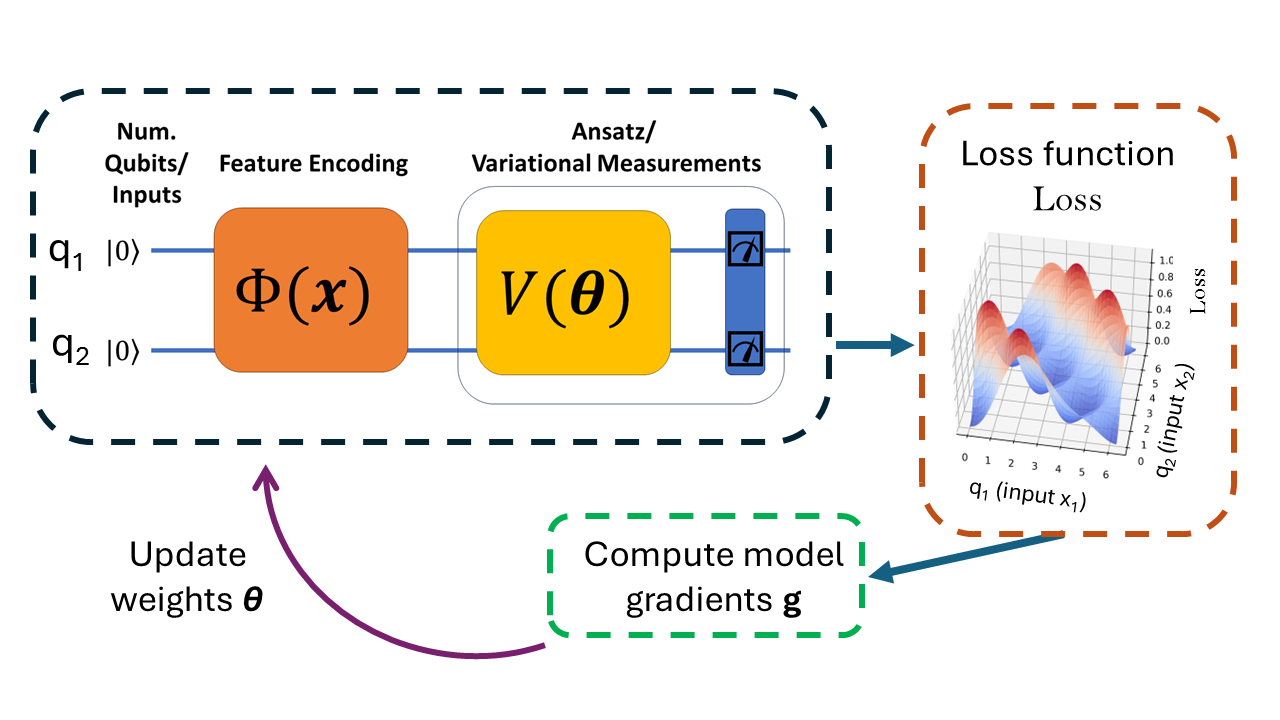}
    \caption{Generic architecture of a VQNN. It consists of a quantum feature map $\Phi(\mathbf{x})$ that encodes classical input data $\mathbf{x}$ into a quantum state. The horizontal lines represent the qubits in the quantum circuit. The ansatz $V(\boldsymbol{\theta})$ is designed with trainable parameters  $\boldsymbol{\theta}$  to explore the solution space. The trainable parameters are optimized by adjusting $\boldsymbol{\theta}$ to minimize a defined loss function ($\mathrm{Loss}$).}
    \label{fig:qnn}
\end{figure}

 VQNNs encode classical data into quantum states and process them using quantum mechanical gates arranged in layers (Fig. \ref{fig:qnn}). The output from measuring the quantum state is then used to calculate the loss function. The trainable parameters are represented as vectors and optimized by measuring gradients using the parameter shift rule \cite{Schuld2019-vo} at each training round. The model parameters are then updated through classical optimization in the training round. VQNN models are flexible and are often combined within larger hybrid quantum-classical models~\cite{Bischof2025}. This allows studying their potential applicability across various domains, including ab-initio molecular modeling~\cite{Cerezo2022}, material analysis\cite{Guan_2021}, high-energy matter \cite{Hirai2024}; and also being explored for general data applications, recognition with ML models, financial applications, and more~\cite{D2CS00203E}.   

\begin{figure}[!t]
    \centering
    \includegraphics[width=0.48\textwidth]{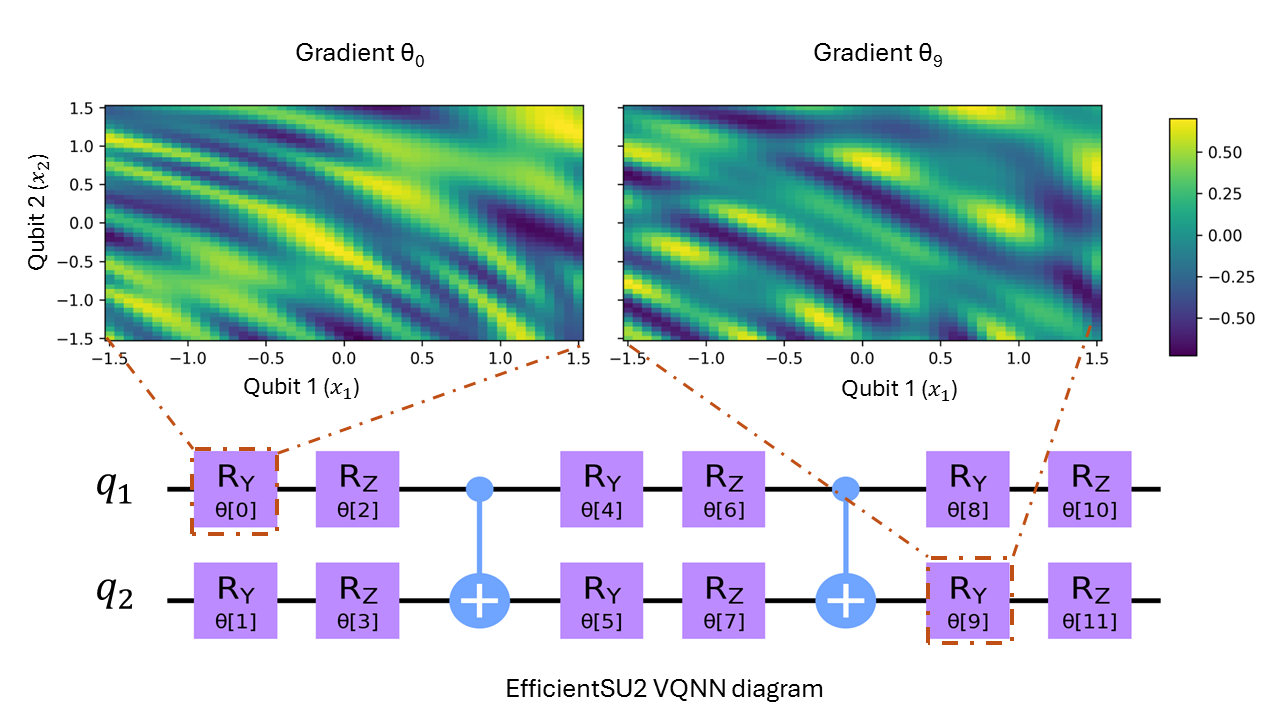}
    \caption{The surface shows the value of the VQNN model loss gradients for each combination of input values. The ansatz VQNN model is used also in our experiments, $EfficientSU2$ \cite{arxiv.2402.16465} with Fraud data as described in Table \ref{tab:experiments} and with input parameters $x_{1}$ (qubit 1) and $x_{2}$ (qubit 2). $ZZFeatureMap$ is used here for embeddings \cite{arxiv.2207.11449, arxiv.2408.10274}.}
    \label{fig:gradients}
\end{figure}

Training VQNNs may prove challenging since gradient-based training suffers from the vanishing gradient problem \cite{McClean2018,Larocca2025} due to barren plateaus in the loss landscape. Another challenge is that the loss landscape is characterized by a large number of local minima. The importance of the latter is apparent even in a system with a few qubits, yet it is somewhat under-looked in the literature (but see \cite{PhysRevResearchlocal}). For instance, Fig. \ref{fig:gradients} shows a sampled gradient map of a two qubits VQNN circuit. This oscillating landscape can lead to large instability in common gradient descent optimizers \cite{PhysRevResearchlocal} (ADAM\cite{kingma2017adammethodstochasticoptimization}, L-BFGS\cite{10.5555/3112655.3112866}, NatGrad\cite{6790500}). Interestingly, the existence of many local minima can be leveraged for privacy-preserving training of VQNN models\cite{kumar2023expressivevariationalquantumcircuits,PhysRevResearch.6.023020,10821342} in distributed learning setups such as \emph{Federated Learning} (FL) \cite{pmlr-v54-mcmahan17a}.
 
\begin{figure}[!t]
    \centering
    \includegraphics[width=0.48\textwidth]{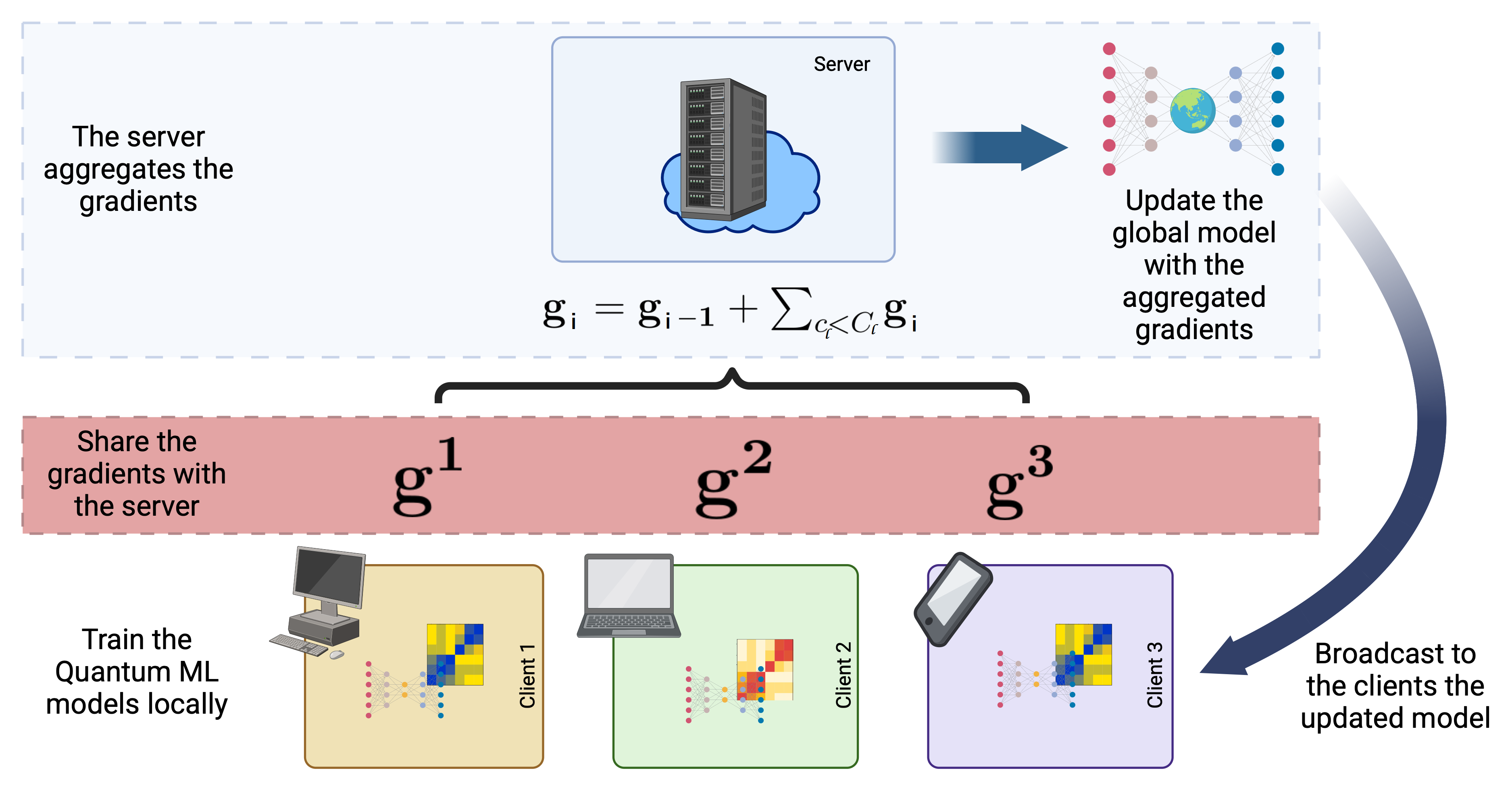} 
    \caption{In a standard pipeline of a Federated Learning model, clients ($C_l$) locally train the neural network on their data and share only the gradients or weights with the server, without exposing the data itself. Then the server aggregates the gradients ($\mathbf{g}$) or weights ($\boldsymbol{\theta}$) and calibrates a global model. Next, it broadcasts the updated global model back to the clients.}
\label{fig:fl-schema}
\end{figure}

Federated learning \cite{pmlr-v54-mcmahan17a} are distributed training paradigms for a machine learning model, without sharing the local data among different parties/clients ($C_l$). FL is crucial in training on sensitive and personal data of edge devices, in communication systems such as in automotive, data-market and pricing data models~\cite{9374643, 9529467, 10.1145/3677127}, or recently, even being used to preserve confidentiality of drug molecule structures in an FL setup between pharmaceutical companies~\cite{Callaway_2025}. 
In a typical application, a central server initializes and distributes a global model to all clients. Each client then independently trains the model on local data, computing weight updates (see Fig. \ref{fig:fl-schema}). Clients send only these updates back to the server. The updated weights $\boldsymbol{\Theta}_{i}$, shared by each client for the model $f$ (with $i$ as the training steps/rounds), are defined as:

\begin{equation}\label{eq:grads}
    \mathbf{G}_{i} = - \frac{\gamma}{B} \sum\limits_{b<B}\frac{\partial \mathrm{Loss} (f(\mathbf{X}_{i,b}, \boldsymbol{\Theta}_{i}), \mathbf{Y}_{i,b})}{\partial \boldsymbol{\Theta}_{i}},
\end{equation}

\noindent
 where $\mathbf{G}_{i}$ the gradients of each training step $i=\{1,2,\cdots, I\}$ with $I$ the total steps, $\mathbf{X}_i$ the features data, $B$ the batch size, $\mathbf{Y}_i$ target values, $\mathrm{Loss}$ the loss function, $\boldsymbol{\Theta}_t$ the model $f$ weights, $\gamma$ the learning rate. The server then aggregates updates of the weights from clients \cite{9599369,QI2024272}, $\mathbf{G}_{i} =\mathbf{G}_{i-1}+\sum_{c_{l}<C_l}{\mathbf{G}_{i}}$, and repeat the training with new rounds of data and possibly new clients. For later use, we define $\mathbf{g}=\mathbf{G}_{i}$ for the gradients, $\boldsymbol{\theta}=\boldsymbol{\Theta}_i$ for the model weights, $\mathbf{x} = \mathbf{X}_i$ for the data, $\mathbf{y} = \mathbf{Y}_i$ for the target values. 

\begin{figure*}[!t]
    \centering
    \subfloat[]{%
        \includegraphics[width=0.57\textwidth]{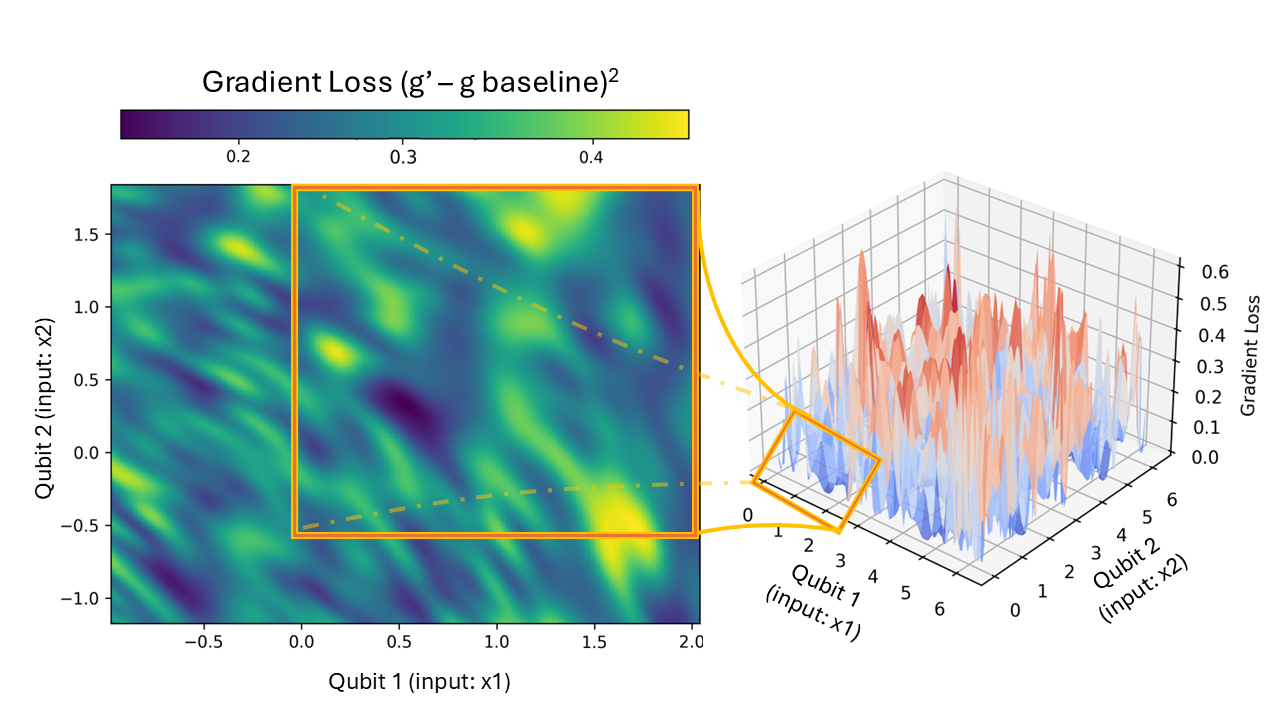}
    }
    \hfill
    \subfloat[]{%
        \includegraphics[width=0.40\textwidth]{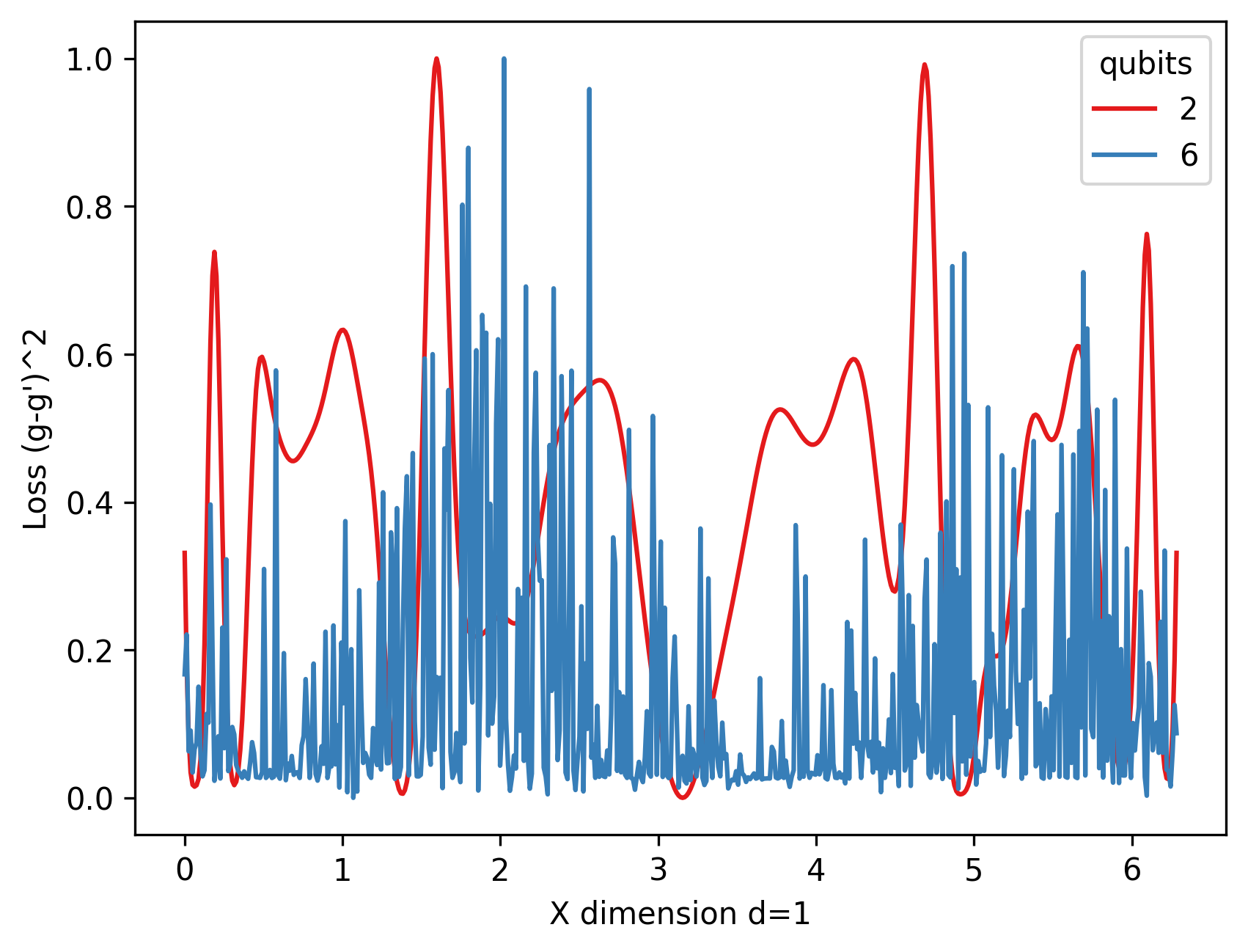}
    }
    \caption{The plots show the gradient loss landscape (Eq.\ref{eq:grad_cost_fun}) with multiple local minima, which can hinder the success of gradient inversion attacks \cite{kumar2023expressivevariationalquantumcircuits,PhysRevResearch.6.023020,10821342}. Each plot shows a different viewpoint of the multiple local minima of a VQNN model. In the 3-D and surface plots ($a$), the $x$-axis and $y$-axis represent the input parameters $x_1$ (qubit 1) and $x_2$ (qubit 2), ranging from 0 to $2\pi$. The $z$-axis depicts the gradient loss $(\mathbf{g}' - \mathbf{g})^2$. The line plot ($b$) shows the multiple local minima across the dimension $x_1$. The VQNN model used in these plots is named `Complex' and depicted in Table \ref{tab:experiments}. }
    \label{fig:3d-landscape}
\end{figure*} 

Gradient sharing is vulnerable to inversion attacks~\cite{lyu2020, 9069945, MOTHUKURI2021619}, allowing adversaries to reconstruct sensitive data from shared gradients. A typical attack involves selecting a proxy model and proxy gradients $\mathbf{g}'$ to minimize a loss function \cite{zhu2019deepleakagegradients, Geiping2020InvertingG, Eloul_2024_WACV} between $\mathbf{g}'$ and $\mathbf{g}$, for example, $||\mathbf{g}' - \mathbf{g}||$ w.r.t a guess proxy input vector $\mathbf{x}'$ \cite{zhu2019deepleakagegradients, Eloul_2024_WACV}. Variations include using vector products instead of absolute distance \cite{10.5555/3495724.3497145}, adding regularizations \cite{Yin2021SeeTG}, or prior distributions to enhance inversion success. Results indicate that NNs are surprisingly highly vulnerable to such attack, with potentially recovering the full large batch of input data within machine-level precision \cite{Zhang_2020_CVPR, s21113874, 9895303}. Proposed mechanisms with Differential Privacy (DP) and other techniques \cite{lyu2020, Eloul_2024_WACV, 10.1145/2810103.2813677} to reduce NN vulnerability often lower model performance, or incur high computational costs with encryption schemes \cite{254465, 9812492, int22818}. Recent studies~\cite{kumar2023expressivevariationalquantumcircuits, Li_2024, 9763352, e23040460} highlight Variational Quantum Circuits (VQCs) as resistant to gradient inversion attacks. VQC gradients, with complex multivariate Chebyshev polynomials, create a system of equations difficult to solve numerically, hindering data reconstruction. Fig.~\ref{fig:3d-landscape} illustrates the challenge of finding a global minimum for $min||\mathbf{g}' - \mathbf{g}||$.

\begin{figure}[!t] 
    \centering 
    \includegraphics[width=0.48\textwidth]{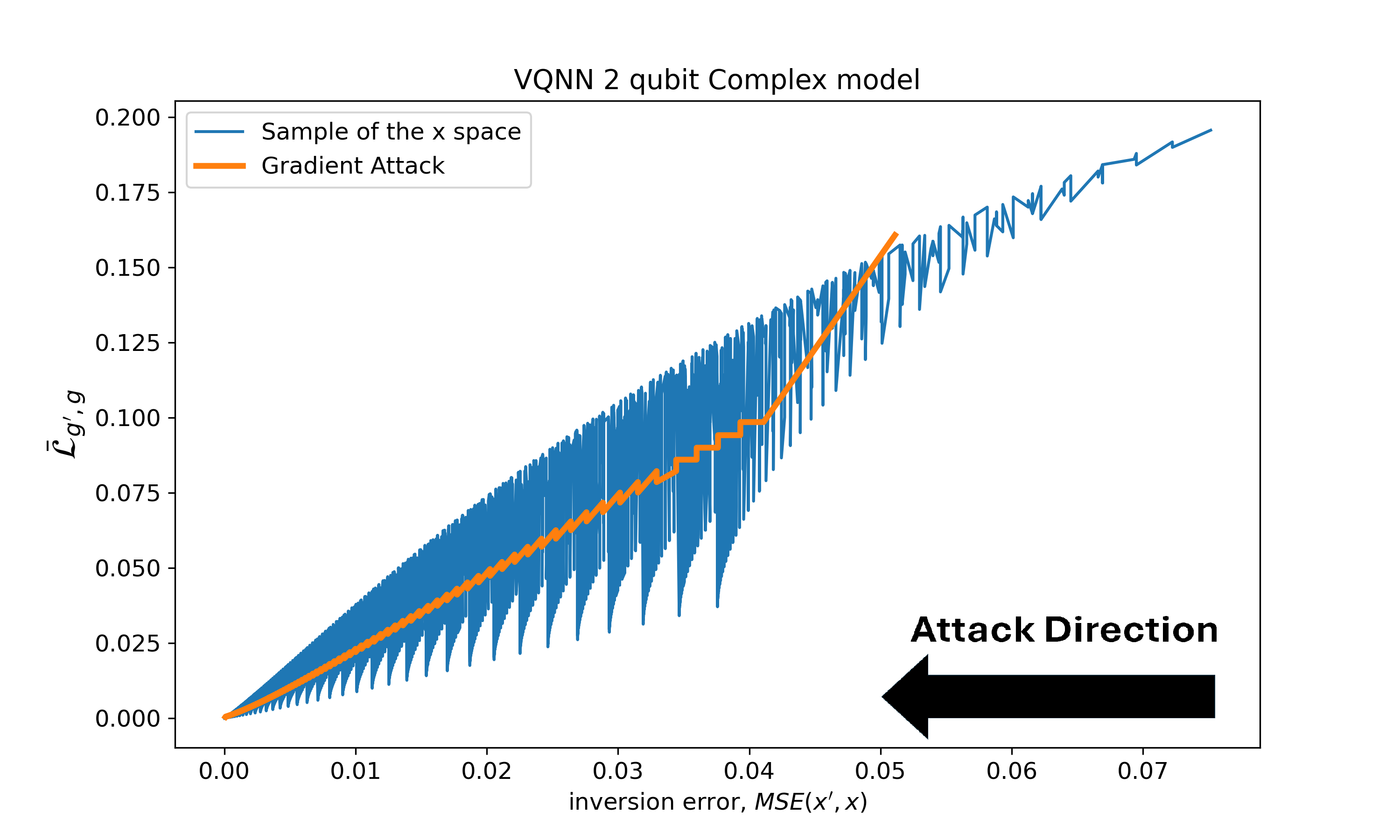} 
    \caption{Gradient inversion attack profile on two qubits `Complex' VQNN model (thick orange line), shows a close-up near the global minimum of the inversion error with $log10$ MSE distance ($\mathbf{x}', \mathbf{x}$) versus the squared gradient loss $(\mathbf{g}' - \mathbf{g})^2$. As the attack is progressed, $\mathbf{g}'$ approaches $\mathbf{g}$ which leads to reveal the hidden $\mathbf{x}$. The fluctuating $\mathbf{x}$ space is plotted (thin blue line) by sampling small random differences of  $(\mathbf{x}'-\mathbf{x})^{2}$. } 
    \label{fig:surface-attack} 
\end{figure}

In this paper we present a new algorithm that successfully adapts inversion attacks to invert private data when training VQNNs in federated or decenteralized learning setups. One could try inversion attacks in the VQNN scheme by numerically calculating gradients using the Finite Difference Method (FDM). However, this alone, would not be sufficient and fail to minimize $min||\mathbf{g}' - \mathbf{g}||$ to the global minimum of the original data, due to the high density of local minima. By incorporating adaptive low-pass filtering in the FDM, the attack can find the global minimum effectively by adapting the low-pass filter's window to the local minima frequency  (see Fig. \ref{fig:surface-attack}). The algorithm is found to be effective as long as the model is trainable. This paper accurately reconstructs input $\mathbf{x}$ with a numerical error of $MSE(\mathbf{x}', \mathbf{x}) \sim 1\mathrm{e}{-10}$. Additionally, when adding a Kalman filter updates, we improve the efficiency of convergence across all trainable models and data, even recovering batch inputs.

\subsection{Contributions}
For a comprehensive privacy study, it is critical to assess the performance of the VQNN models. Therefore, this work revisits training tasks proposed on VQNNs. It is shown, that practical models which are trainable with typical optimizer (SGD or L-BFGS), can be also vulnerable to the inversion algorithm proposed here. In addition, the work includes results for benchmark dataset (MNIST \cite{Lecun726791}) and importantly, real-life sensitive data (Credit Card Fraud \cite{ULB_2018}). Through this systematic study, we provide a practical analysis of data leakage risks in sensitive QML applications. Although our work provides results on representative and typical cases, it does not intend to provide a full assessment to all potential models or datasets. Instead, it forms the base for a benchmark scheme that can be tested in a specific application or for further research on security of model training. 

Specifically, our contributions are:
\begin{itemize}
  \item \textbf{A successful gradient inversion attack on VQNN models. The attack is achieved by an optimization algorithm based on finite difference gradient estimation with an adaptive filtering.}
  \item \textbf{A critical study of the gradients inversion in trainable VQNN models, at different sizes and complexity, batching, and various datasets.}
  \item \textbf{Providing a practical benchmark attack to assess the trade-off between privacy and performance, and in comparison to NN models.}
\end{itemize}

The rest of the paper is structured as follows: Section \ref{sec:literature} overviews recent work on privacy in QML followed by background on VQNN model training (Section \ref{sec:background}). Section \ref{sec:methodology}, details the method for the inversion attack with the proposed optimization algorithm (Alg. \ref{alg:batch}). Section \ref{sec:experiments} discusses the VQNN models used, their training and setup (Table \ref{tab:experiments}), assumptions, and the experiments carried out. Then, we discuss results of inversion attacks in the various experiments of models, datasets, and batch. We also compare performance versus privacy,  with differential privacy using noisy gradients in VQNN and NN (Section \ref{sec:results}).

\section{Related Work}
\label{sec:literature}
Many recent works explored the applicability of quantum FL setup and indicated the added value for privacy \cite{e23040460, Li_2021, 9763352, kumar2023expressivevariationalquantumcircuits, heredge2024prospectsprivacyadvantagequantum, Li_2024}. However, none of these works demonstrate a successful attack for practical cases, and hence our paper provides an important insight to practitioners and researchers in the field. \cite{Yang_2021} uses a hybrid quantum convolutional network to preserve privacy via FL for automatic speech recognition. \cite{Watkins2023} focuses on implementing privacy-preserving QML through the application of DP. The authors propose a hybrid quantum-classical model using a VQC trained with differentially private optimization algorithms. The privacy mechanism is introduced by clipping the gradients and adding Gaussian noise to ensure DP. Compared to their work, our paper goes beyond simple noise addition for privacy by actively exploring vulnerabilities through gradient-based attacks, specifically gradient inversion. Other works in quantum DP \cite{10115324} which primarily focus a theoretical framework for quantum DP using information-theoretic tools like the hockey-stick divergence. Their work does not actively explore attacks on quantum models to evaluate privacy risks.  
\cite{Rofougaran2024} employ quantum FL, which distributes training data among multiple clients and trains local models iteratively. They also integrate quantum DP by adding noise during training. However, they do not explicitly perform adversarial or propose inversion attacks to reconstruct original inputs. Their emphasis is on accuracy versus privacy under FL. 

Closely related works are in a new line of research, \cite{kumar2023expressivevariationalquantumcircuits,icaart24}. These works investigate properties of `overparameterized' QML models with highly-expressive encoding (in terms of Fourier frequency spectrum) that provide an inherent protection against gradient inversion attacks. It examines these models in the context of an FL setup. The works show theoretical results backed by empirical evidence that the perpetrator encountered a considerable challenge when attempting to craft an attack on these models due to its complexity. Here, we show empirically, that despite this challenge, we can successfully invert to original data in trainable VQNN models. \cite{heredge2024prospectsprivacyadvantagequantum} establishes a relation between the privacy of VQNNs and the dimension of the Lie algebra of their circuit generators. This theoretical insight highlights the general susceptibility of various models to analytical closed-form solution attacks. Our research complements this study and introduces the first algorithm for inversion attacks of VQNNs. This is valuable tool for exploring and benchmarking the resilience of VQNNs in practical applications.
Several recent works also discuss the security and privacy of VQNNs \cite{franco2024} mostly with relation to adversarial attacks (or white-box attacks since they require access to target values) \cite{Biggio2013, arxiv.1608.04644}, a group of methods that exploit gradients to create inputs (adversarial examples) that fool the model into making incorrect predictions. For example, 
\cite{Liao2021} and \cite{Weber2021} explore adversarial robustness in VQNNs. \cite{Liao2021} analyzes vulnerabilities under scenarios involving Haar-random pure states and Gaussian latent spaces, using the quantum DP framework from \cite{Zhou2017}. They provide robustness bounds and focus on classification tasks, particularly examining prediction-change attacks and their impact on artificial datasets like MNIST \cite{Weber2021}, on the other hand, leverage quantum hypothesis testing (QHT) to establish formal robustness conditions for quantum classifiers, proposing assessments to evaluate resilience against adversarial and noise perturbations. Both works emphasize robustness evaluation not on the inversion attack type presented here, and require target labels or classification outputs.

In contrast, our work departs from these frameworks by implementing gradient-based inversion attacks that exploit quantum reversibility without relying on target labels or classification outputs. By approximating original input data in both regression and classification contexts, we provide a broader vulnerability analysis encompassing both benchmark and real-world datasets, revealing additional risks not covered in prior studies. Similar to \cite{Li_2021,Liao2021,Weber2021}, \cite{PhysRevResearch.3.023153} investigate adversarial robustness in VQNNs by incorporating quantum noise (similar to DP), proposing it as a defense mechanism against adversarial attacks. Their work employs the iterative fast gradient sign method (I-FGSM) to craft adversarial examples, which rely on access to target labels for calculating perturbations that maximize the model’s loss. This approach evaluates robustness by measuring the rate of mis-classification caused by these perturbations, with a focus on assessing the effects of noise on classification accuracy. However, their methodology is inherently dependent on the availability of target values and is designed for classification tasks. In contrast, our work operates without access to target labels, employing gradient inversion techniques that exploit quantum reversibility to recover input data directly.

\section{Background}
\label{sec:background}
\subsection{Training Variational Quantum Neural Networks (VQNN)}
\label{sec:quant_ml}
Given a dataset $\{(\mathbf{X}_i, \mathbf{Y}_i)\}_{i=1}^I$ that consists of $\mathbf{X}_i \in \mathbb{R}^J$ input features and corresponding target values for classification $\mathbf{Y}_i \in \{ 0, 1 \}^C$ where $C$ is the number of classes or in regression models $\mathbf{Y}_i \in \mathbb{R}$. In a model the parameters are $\boldsymbol{\theta}=\{\theta_1, \theta_2, \cdots, \theta_m, \cdots, \theta_M\}$, with $m=[1,M]$. Subsequently, training a neural network with parameter $\boldsymbol{\theta}$ that accepts the data $\mathbf{X}_i$  as input, it requires minimizing a loss function between the prediction $\mathbf{\hat{Y}}_i = f(\mathbf{X}_i, \boldsymbol{\theta})$ and the target values $\mathbf{Y}_i$:

\begin{equation}
    \min_{\boldsymbol{\theta}}\left[\sum_{i=1}^{I} \mathrm{Loss}(\mathbf{\hat{Y}}_i, \mathbf{Y}_i)\right]
\end{equation}

In order to minimize the loss, the gradients (Eq.\ref{eq:grads}) of the loss function with respect to the model parameters $\boldsymbol{\theta}$ are calculated. After the gradient estimation, the model parameters can be updated by optimization such as stochastic gradient descent update to complete one round of training. In VQNN, the computation of gradients is different and requires the measurement of the qubits' states as detailed below.

\begin{algorithm}[!t]
\caption{Training Algorithm for VQNN}\label{alg:train_qnn}
\label{alg:training_vqnn}
\begin{algorithmic}[1]
\Require Dataset $\mathbf{x}$ with rows $\{(\mathbf{X}_i, \mathbf{Y}_i)\}_{i=1}^I$, initial parameters $\boldsymbol{\theta}_0$, learning rate $\eta$, maximum epochs $I$
\Ensure Optimized parameters $\boldsymbol{\theta}^*$
\State Initialize $\boldsymbol{\theta}^* \gets \boldsymbol{\theta}_0$
\For{$i = 1$ to $I$}
    \For{each $(\mathbf{X}_i, \mathbf{Y}_i)$ in the dataset  $\mathbf{x}$ }
        \State Encode $\mathbf{X}_i$ into quantum state:
        \State $|\Phi(\mathbf{X}_i)\rangle \gets U_\Phi(\mathbf{X}_i)|0\rangle^{\otimes q}$
        \State Apply parameterized circuit: 
        \State $|\psi(\mathbf{X}_i, \boldsymbol{\theta}^*)\rangle \gets V(\boldsymbol{\theta}^*)|\Phi(\mathbf{X}_i)\rangle$
        \State Measure observable: 
        \State $\hat{\mathbf{Y}}_i \gets \langle \psi(\mathbf{X}_i, \boldsymbol{\theta}^*)|\hat{O}|\psi(\mathbf{X}_i, \boldsymbol{\theta}^*) \rangle$
        \State Compute loss: 
        \State $ \mathrm{Loss} \gets ( \hat{\mathbf{Y}}_i - \mathbf{Y}_i)^2$
        \State Compute gradients $\mathbf{g} = $ Eq. \ref{eq:param_shift}
        \State Update parameters: $\boldsymbol{\theta}^* \gets \boldsymbol{\theta}^* - \eta \mathbf{g}$
    \EndFor
\EndFor
\State \Return $\boldsymbol{\theta}^*$
\end{algorithmic}
\end{algorithm}

Alg. \ref{alg:train_qnn} summarizes the training of a VQNN model. As a first step, it uses \emph{quantum feature map} $\Phi(\mathbf{x})$ to encode the classical input data $\mathbf{x}$ into a quantum state \cite{McClean2018,Cerezo2021}. 

\begin{equation}\label{eq:featuremap}
|\Phi(\mathbf{x})\rangle = U_{\Phi}(\mathbf{x}) |0\rangle^{\otimes q}, 
\end{equation}

where $U_{\Phi}(\mathbf{x})$ is a unitary operator parameterized by $\mathbf{x}$, and $q$ is the number of qubits. The feature map $\Phi(\mathbf{x})$ is a circuit of quantum gates that transforms the initial state $|0\rangle^{\otimes q}$ with a unitary operation $U_{\Phi}$ that is embedded with the information of $\mathbf{x}$. Typically, in angle encoding feature maps, the number of qubits is chosen to be equal to the number of dimensions in $\mathbf{x}$, so $J=q$, where $\mathbf{x}=\{x_1, x_2, \cdots, x_J\}$ and $j$ the dimension index $j = \{1, 2, \cdots, J\}$. A feature map used in the models here is a rotation gate around the X-axis ($R_{\text{X}}$)\cite{McClean2018,Cerezo2021}, with $R_{\text{X}}(x_j) = \exp\left(-i \frac{x_{j}}{2} \sigma_x\right)$ and $\sigma_x$ is the Pauli-X matrix. The \emph{ansatz} $V(\boldsymbol{\theta})$ is a parameterized quantum circuit that consists of sequence of quantum gates with trainable parameters $\boldsymbol{\theta}$ to create a highly expressive model\cite{das2023}:

\begin{equation}\label{eq:ansatz}
|\psi(\mathbf{x}, \boldsymbol{\theta})\rangle = V(\boldsymbol{\theta}) |\Phi(\mathbf{x})\rangle
\end{equation}

the ansatz includes rotations around the Y and Z axes and controlled-NOT (CNOT) gates to entangle qubits. To obtain a prediction, a \emph{measurement} on the quantum state $|\psi(\mathbf{x}, \boldsymbol{\theta})\rangle$ is performed. A pre-defined Hermitian operator is used as an \emph{observable} $\hat{O}$ and the expected value is computed \cite{McClean2018}:

\begin{equation}\label{eq:measurement}
\hat{\mathbf{y}} = \langle \psi(\mathbf{x}, \boldsymbol{\theta}) | \hat{O} | \psi(\mathbf{x}, \boldsymbol{\theta}) \rangle.
\end{equation}

A common choice of the observable is the Pauli-Z ($Z$) operator $\sigma_z$, but any other operator can be chosen based on the problem. In classification tasks we encode probabilities with the Pauli-Z operator to measure the state of each qubit. From Eq. \ref{eq:measurement}, for regression tasks the $\hat{\mathbf{y}}$ is a continuous value derived from the expectation value of the observable. To train a VQNN, the parameters are measured and gradients are estimated with the \emph{parameter-shift rule}~\cite{Li2017-qv, Mitarai2018-dx, Schuld2019-vo, Wierichs2022-pk}:

\begin{equation}\label{eq:param_shift}
g_m = \frac{1}{2} \left[ \hat{\mathbf{y}} \left(\boldsymbol{\theta}  + \frac{\pi}{2} \hat{\mathbf{e}}_m \right) - \hat{\mathbf{y}} \left(\boldsymbol{\theta} - \frac{\pi}{2} \hat{\mathbf{e}}_m \right) \right]
\end{equation}

where $\hat{\mathbf{e}}_m$ is the unit vector of size $M$, the $m$-th entry equals one $[\hat{\mathbf{e}}_m]_m = 1$, and all other entries $[\hat{\mathbf{e}}_m]_{i \neq m} = 0$. After evaluation of the gradients, the model parameters can be optimized with similar methods to classical NN.

\section{Methods}
\label{sec:methodology}

\subsection{Gradient Inversion Attacks}
\label{sec:gradient_attacks}

The core idea behind gradient inversion attacks is to exploit the parameters shared during or after a distributed training. An adversary  tries to infer the original input data by analyzing these gradients. For example, by using the `Deep Leakage Gradient' (DLG) attack on classical NN~\cite{zhu2019deepleakagegradients}. In classical NNs, the optimization of $\mathbf{x}'$ to recover $\mathbf{x}$  relies  simply on backpropagation of a differentiable model to calculate some cost function between $\mathbf{g}'$ and $\mathbf{g}$, denoted by $\mathcal{L}_{\mathbf{g}', \mathbf{g}}$, and differentiate it with respect to $x'_j$,  $ \frac{\mathcal{L}_{\mathbf{g}', \mathbf{g}}}{\partial {x_j'}} $. Then, to use an optimizer to find $\mathbf{x}'$ that minimizes the cost function~\cite{Huang2020InstaHideIS,Yin2021SeeTG}, or even invert it directly for some NN architectures~\cite{Eloul_2024_WACV}. In a VQNN, the gradients are obtained with a measurement, hence we adapt the finite-difference method to estimate numerically $\frac{\partial  \mathcal{L}_{\mathbf{g}', \mathbf{g}}}{\partial x_{j}}$ for every $j$. However, the main challenge, is to optimize $\mathcal{L}_{\mathbf{g}', \mathbf{g}}$ to the global minimum. This work proposes an adaptive low-pass filter in the finite difference estimation of gradients. The cost function used here, is the euclidean distance between the gradients $\mathbf{g}'$ and $\mathbf{g}$, where $\mathbf{g}'$ is the gradient of Eq. \ref{eq:grads} with  proxy model input using $\mathbf{x}'$:

\begin{equation}\label{eq:grad_cost_fun}
\mathcal{L}_{\mathbf{g}', \mathbf{g}} = \frac{1}{M} \sum_{m=1}^{M} \left( g_m' - g_m \right)^2
\end{equation}
Other cost functions and regularization can also be considered~\cite{Zhang_2020_CVPR}, but are beyond the scope of the study.
The inversion attack scheme has the following steps:

\textbf{1. Numerical Gradient Approximation}:
    For each dimension $ j $, compute the numerical gradient of the cost function with respect to $ x_j $ (the $j$ similar to Eq. \ref{eq:featuremap}):

    \begin{equation}\label{eq:grad_cost_approx}
    \frac{\partial \mathcal{L}_{\mathbf{g}', \mathbf{g}}^{n}}{\partial x'_j} = \frac{\mathcal{L}_{\mathbf{g}', \mathbf{g}}(x'_j + nhe_j) -\mathcal{L}_{\mathbf{g}', \mathbf{g}}(x'_j - nhe_j)}{2nh}+O(nh)
    \end{equation}
     See Alg. \ref{alg:batch}, and $ e_j $ is the unit vector in the $j$-th direction. $n$ the moving average window size used in the algorithm $n=[1,N]$. $h$ the step size of the algorithm.

\textbf{2. Low-pass filter}:
    We can use a simple moving average window with size $N$, to remove oscillations with frequencies,  $1/(Nh)$:
       \begin{equation}\label{eq:dldx}
    \frac{\partial \mathcal{L}_{\mathbf{g}', \mathbf{g}}}{\partial x'_j} =\frac{1}{N}\sum_{n=1}^N\frac{\partial \mathcal{L}_{\mathbf{g}', \mathbf{g}}^{n}}{\partial x'_j}+O(nh) 
    \end{equation}
    The window size for filtering,
    $Nh$ depends on the model complexity and how close $\mathcal{L}_{\mathbf{g}', \mathbf{g}}$ is to zero, the global minima. $Nh$ is initially chosen to be sufficiently large and is dynamically adapted according to convergent criteria. Optimally, close to the global minima, reducing $Nh$ as $\mathcal{L}_{\mathbf{g}', \mathbf{g}}|_{Nh\rightarrow0}\rightarrow0$.

\textbf{3. Gradient Descent Update}:
    It is sufficient to use a stochastic gradient descent where at each round, $k$, each dimension of $ \mathbf{x}' $ is updated:
    \begin{equation}
    {x'_j}^{k} \leftarrow {x'_j}^{k-1} - lr \frac{\partial \mathcal{L}_{\mathbf{g}', \mathbf{g}}}{\partial x'_j}
    \label{sgd}
    \end{equation}
    where $ lr $ is the learning rate, obtaining the updated $\mathbf{x}'^k$ from the previous $\mathbf{x}'^{k-1}$.

\textbf{4. Iterative Refinement}:
    Repeat the gradient approximation and update steps iteratively for a fixed number of iterations ($K$) or until reaching convergence. 
    Implementation of the steps are presented in Alg. \ref{alg:batch}.

\begin{algorithm} 
\caption{Evaluate Gradient Inversion Susceptibility in VQNNs with Kalman Filter}\label{alg:batch} 
\begin{algorithmic}[1] 
\Require $\mathbf{x}'$ (initial random data), $N$ (moving average window size), $h$ (step size), $lr$ (learning rate), $J$ (number of dimensions $\mathbf{x}'$), $\mathbf{g}$ (gradient of original data $\mathbf{x}$), $\text{threshold}$ (gradient change), $K$ the number of iterations
\Ensure Optimized $\mathbf{g}'$ approaching $\mathbf{g}$ 
\State $\mathbf{g} \gets$ VQNN gradients of original data $\mathbf{x}$ from clients $C$ 
\State Initialize ($k=0$) Kalman filter parameters as all-ones elements ($\mathbf{F}, \mathbf{H}$) and zero initial state $\mathbf{x}'^{k-1}$, see Appendix, Eq. \ref{eq:kalman_predict} and \ref{eq:kalman_update}.
\For{$k = 1$ to $K$}
\For{$j = 1$ to $J$} 
    \State Initialize $loss_{+} \gets 0$ and $loss_{-} \gets 0$ 
    \For{$n = 1$ to $N$} 
        \State $\mathbf{x}'_{+} \gets \mathbf{x}'$ 
        \State $x'_{+, j} \gets x'_{+, j} + (n+1) \cdot h$ 
        \State $\mathbf{g}'_{+} \gets$ VQNN parameter-shift on $\mathbf{x}'_{+}$ (Eq. \ref{eq:param_shift})
        \State $loss_{+} \gets loss_{+} + \mathcal{L}_{\mathbf{g}'_{+}, \mathbf{g}}$  (Eq. \ref{eq:grad_cost_fun})

        \State $\mathbf{x}'_{-} \gets \mathbf{x}'$
        \State $x'_{-,j}\gets x'_{-, j} - (n+1) \cdot h$
        \State $\mathbf{g}'_{-} \gets$ VQNN parameter-shift on $x'_{-}$ (Eq. \ref{eq:param_shift})
        \State $loss_{-} \gets loss_{-} + \mathcal{L}_{\mathbf{g}'_{-}, \mathbf{g}}$  (Eq. \ref{eq:grad_cost_fun})
        \State Numerical gradient approximation (Eq. \ref{eq:grad_cost_approx}):
        \State $ \frac{\partial \mathcal{L}_{\mathbf{g}', \mathbf{g}}^{n}}{\partial x'_j} \gets (loss_{+} - loss_{-}) / (2 \cdot n \cdot h)$
    \EndFor
    \State Numerical gradient approximation (Eq. \ref{eq:dldx}):
    \State $ \frac{\partial \mathcal{L}_{\mathbf{g}',\mathbf{g}}}{\partial x'_j}$
    \State Compute average $\mathcal{\bar{L}}_{\mathbf{g}',\mathbf{g}}^{k}$:
    \State $\mathcal{\bar{L}}_{\mathbf{g}',\mathbf{g}}^{k} \gets \frac{1}{N}\sum^{2N} (loss_{+} + loss_{-})$
    \State ${x'_j}^{k} \gets {x'_j}^{k-1} - lr \cdot  \frac{\partial \mathcal{L}_{\mathbf{g}', \mathbf{g}}}{\partial x'_j}$
    \State ${x'_j}^{k} \gets$ Perform Kalman update (Eq. \ref{eq:kalman})
\EndFor

    \If{$(\mathcal{\bar{L}}_{\mathbf{g}',\mathbf{g}}^{k}/\mathcal{\bar{L}}_{\mathbf{g}',\mathbf{g}}^{k-1}) > \text{threshold}$}
        \State $N \gets \max(1, N // 2)$ 
    \EndIf
\EndFor 
\State \Return $\mathbf{x}'$ \end{algorithmic} 
\end{algorithm}

\begin{table*}[!t]
\caption{VQNN model training performance and architectures. The $ZZEfficient$ model in the experiments is the $ZZFeatureMap$ \cite{arxiv.2207.11449, arxiv.2408.10274} and $EfficientSU2$ \cite{arxiv.2402.16465} models. All models use a $ \mathrm{Loss}$ function of the MSE. The Cosine is a regression model and the Fraud and MNIST are binary classification models. The training/test results are the average of 5-fold Cross Validation. }
\label{tab:experiments}
\resizebox{\textwidth}{!}{%
\begin{tabular}{cccccccc}
\hline
Dataset & Model & Qubits/Inputs & Model Diagram & Metric & Train & Test & Attack Plot \\ \hline
Cosine & Simple & 2 & Fig. \ref{fig:architec-models}\textcolor{blue}{a}    & $R^2$ & 99\% & 96\% & Appx. Fig. \ref{fig:cosine}  \\
 & Simple & 6 & Fig. \ref{fig:architec-models}\textcolor{blue}{a}    &  $R^2$ & 98\% & 94\% & Appx. Fig. \ref{fig:cosine}   \\
 & Complex & 2 & Fig. \ref{fig:architec-models}\textcolor{blue}{c}   &  $R^2$ & 99\% & 98\% & Appx. Fig.  \ref{fig:cosine}   \\
 & Complex & 6 & Fig. \ref{fig:architec-models}\textcolor{blue}{c}    &  $R^2$  & 99\% & 95\% &  Fig. \ref{fig:overview_attacks} \\
MNIST & Simple & 2 & Fig. \ref{fig:architec-models}\textcolor{blue}{a}    & Balanced Acc. & 89\% & 80\% & Appx. Fig. \ref{fig:mnist} \\
 & Simple & 6 & Fig. \ref{fig:architec-models}\textcolor{blue}{a}    & Balanced Acc. & 91\% & 81\% & Appx. Fig. \ref{fig:mnist}  \\
 & Complex & 2 & Fig. \ref{fig:architec-models}\textcolor{blue}{c}     &  Balanced Acc. & 84\% & 81\% & Appx. Fig. \ref{fig:mnist}  \\
 & Complex & 6 & Fig. \ref{fig:architec-models}\textcolor{blue}{c}    &  Balanced Acc. & 83\% & 80\% & Fig. \ref{fig:overview_attacks} \\
Credit Card Fraud & Simple (Entangled) & 2 & Fig. \ref{fig:architec-models}\textcolor{blue}{b}    &   Balanced Acc. &  88\% & 86\% & Appx. Fig. \ref{fig:fraud}  \\
 & Simple (Entangled) & 6 & Fig. \ref{fig:architec-models}\textcolor{blue}{b}    & Balanced Acc.  &  85\% & 83\% & Appx. Fig. \ref{fig:fraud}  \\
 & Complex & 2 & $ZZEfficient$ (Fig. \ref{fig:architec-models}\textcolor{blue}{d})    &   Balanced Acc. & 84\% & 81\% & Appx. Fig. \ref{fig:fraud}  \\
 & Complex & 6 & $ZZEfficient$ (Fig. \ref{fig:architec-models}\textcolor{blue}{d})  &  Balanced Acc. & 80\% & 78\% & Fig. \ref{fig:overview_attacks}  \\ \hline
\end{tabular}%
}
\end{table*}

\subsection{Kalman Filter Attack}
To enhance the attack's efficiency, we employ a Kalman filter, which iteratively refines the estimate of the input values. The Kalman filter operates in two parts: prediction and measurement. The prediction part uses transition $\mathbf{F}$ matrix to predict new state from previous iteration $\mathbf{x}'^{k-1}$ and initialized to zero in $k=0$. The measurement here is calculated from the new $\mathbf{x}'^{k}$ in Eq. \ref{sgd} to obtain the update:    

\begin{align}\label{eq:kalman}
    \mathbf{x}'^k &= \mathbf{F} \mathbf{x}'^{k-1}+ \mathbf{D}^k(\mathbf{x}'^k - \mathbf{H} \mathbf{F}\mathbf{x}'^{k-1})
\end{align}

$\mathbf{H}$ is the measurement matrix, and $\mathbf{D}^k$ is the Kalman gain matrix for the correction of the prediction. We explain the details of the update stages for obtaining $\mathbf{D}^k$, with a standard application of Kalman filter \cite{10.5555/897831} in Appendix, Eq. \ref{eq:kalman_predict} and \ref{eq:kalman_update}.

\section{Experiments}
\label{sec:experiments}
\subsection{Data Generation and Quantum Model Design}
\begin{figure}[!t]
    \centering
    \includegraphics[width=0.48
    \textwidth]{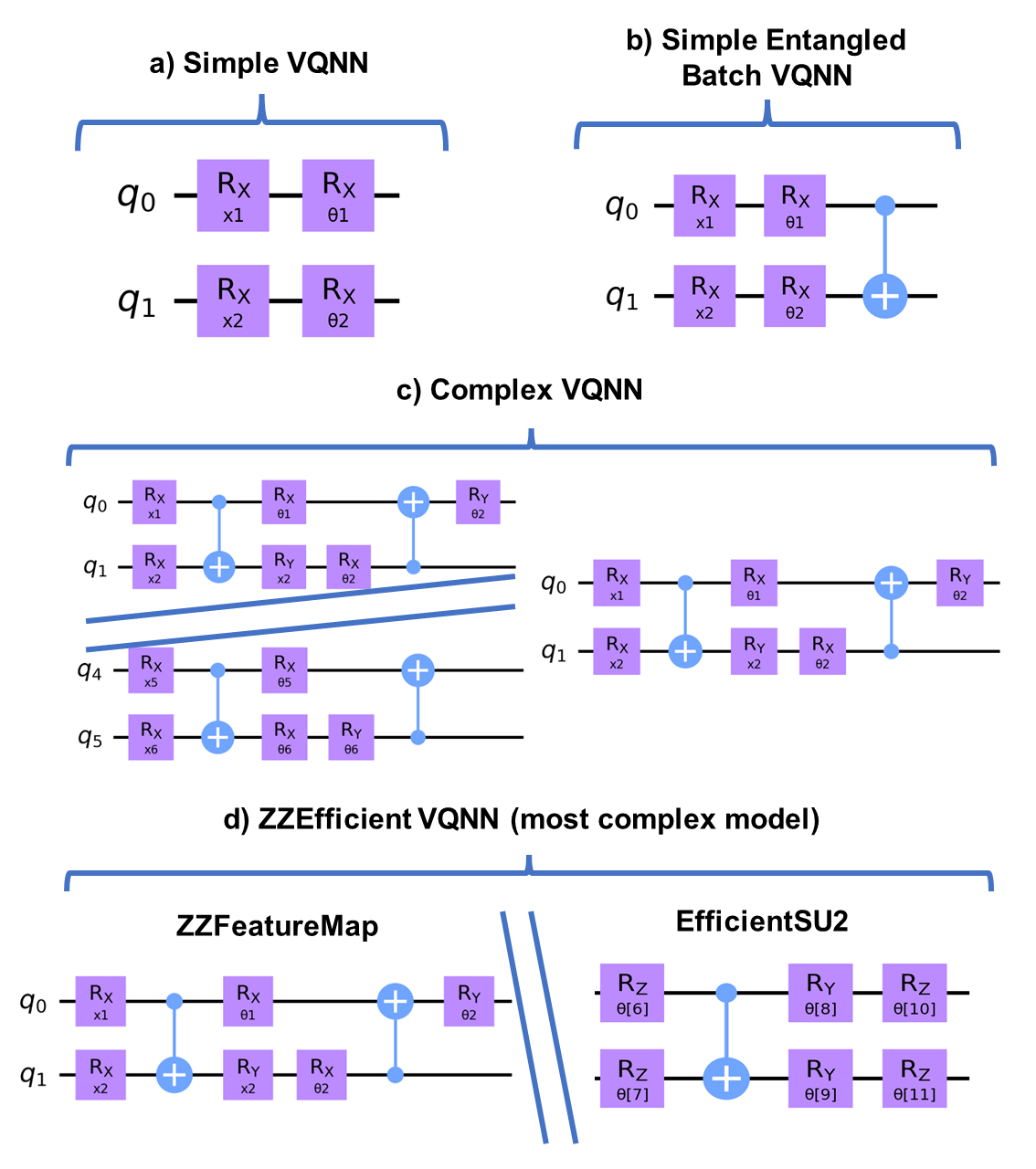} 
    \caption{The model architectures which are used in the experiments. A Simple VQNN system $(a)$ and a Complex designed system $(c)$ for both regression and classification for the generated Cosine and the MNIST datasets. This schematic shows a 2 and a 6-dimensional qubit/input model. System $(b)$ is designed for batch data. For the Credit Card Fraud experiment, the trainable architecture $ZZFeatureMap$ \cite{arxiv.2207.11449, arxiv.2408.10274} and $EfficientSU2$ \cite{arxiv.2402.16465} $(d)$ are used, named as $ZZEfficient$ in the paper.}
\label{fig:architec-models}
\end{figure}

We conduct model training and attack experiments on various benchmark datasets, with different VQNN architectures and qubit/input numbers, for both regression and binary classification tasks. The datasets used are MNIST \cite{Lecun726791}, data generated from a commonly used cosine function with $\mathbf{y} = 0.7 \cos(\mathbf{x})$, and Fraud detection data \cite{ULB_2018}. Model architectures are shown in Fig. \ref{fig:architec-models}, including $ZZFeatureMap$ \cite{arxiv.2207.11449, arxiv.2408.10274} and $EfficientSU2$ \cite{arxiv.2402.16465} (referred here as $\mathbf{ZZEfficient}$). Experiments involve systems of 2 and 6 qubits and inputs. Table \ref{tab:experiments} displays VQNN training performance comparable to classical NNs. It includes 5-fold Cross Validation averages, with datasets split into 80/20. We apply the proposed attack in this study to shared gradients in a round of training. The experiments are carried out on untrained models, but complementary results for trained model are provided in Appendix Fig. \ref{fig:trained_model_attack}. Qiskit \cite{qiskit2024} and Pennylane \cite{pennylane} libraries are both used through out the study with combination of Torch library to evaluate VQNNs model gradients using the parameter shift rule, and for training the models.

\subsection{Inversion attack experiment}
Table \ref{tab:attacks} provides an overview of the attack experiments. The attack algorithm in Section \ref{sec:gradient_attacks} is used here to optimize random initial points $ \mathbf{x}' $ with adaptive moving average and also with Kalman filter. Experiments are carried out on $J=2 $ and $ J=6 $ dimensional datasets, for various different circuit architectures (Fig. \ref{fig:architec-models}).

\subsection{Batch Inversion Attack}
In an additional experiment, we explore a batch attack on a VQNN (Table \ref{tab:attacks}). In this experiment, gradients are summed over a batch of vectors, before being shared. Overall, the larger the batch is, the harder is to invert the gradients to the input vectors. In VQNN, a batch size of 2 and 3 are used. Larger batch sizes causes significant degradation of model performance due to their small model sizes. In this experiment, the model used is a Simple Entangled VQNN depicted in Fig. \ref{fig:architec-models}. The CNOT gate introduces entanglement between qubits, which maximizes the dependencies between the input elements in the model. This ensures an `Over-parameterized' model.

\begin{table*}[!t]
\caption{Attacks on a variety of VQNN models, with single row data and batch-size data. The results are the sample average $MSE(\mathbf{x}',\mathbf{x})$ of 10 attacks with different starting points $\mathbf{x}'$. The $(\pm SE)$ is the standard error of the results. Successful attacks are considered to reach a threshold of $\leq 0.005$ MSE. The model architecture is based on Fig. \ref{fig:architec-models}. Maximum iterations of the models are $\leq 250$.}
\label{tab:attacks}
\resizebox{\textwidth}{!}{
\begin{tabular}{|l|l|l|l|l|}
\hline
 Model & One Shot Success \% & Adaptive Moving Average MSE $(\pm SE)$  & One Shot Success \% & Kalman MSE $(\pm SE)$ \\ \hline
VQNN 2-qubits Cosine Complex  & 100\% & 0.0002 $(\pm 7\mathrm{e}{-4})$  & 100\% & 0.0002 $(\pm 5\mathrm{e}{-4})$ \\ 
VQNN 6-qubits Cosine Complex & 40\% & 0.0005 $(\pm 4\mathrm{e}{-4})$ & 70\% & 0.0003 $(\pm 3\mathrm{e}{-4})$ \\ 
VQNN 6-dims Fraud ZZEfficient & 50\% & 0.0007$(\pm 2\mathrm{e}{-4})$  & 70\% & 0.0003$(\pm 1\mathrm{e}{-4})$ \\ 
VQNN 6-dims MNIST Complex & 40\% & 0.0005$(\pm 1\mathrm{e}{-4})$ & 80\% & 0.0002$(\pm 1\mathrm{e}{-4})$ \\ 
VQNN Batch size 2 & 100\% & 0.00005 $(\pm 1\mathrm{e}{-5})$ & 100\% & 0.00002 $(\pm 4\mathrm{e}{-5})$ \\ 
VQNN Batch size 3  & 100\%  & 0.002 $(\pm 0.005)$ & 100\% &  0.003 $(\pm 0.004)$ \\ \hline
\end{tabular}
}
\end{table*}

\subsection{Differential Privacy and NN Experiment}
Experiments for benchmark NN models carried out to compare both training, and  privacy against inversion attack, under differential privacy schemes involving adding noise. A Gaussian noise with  standard deviation $\sigma$, is added to the shared original gradients $\mathbf{g}$. The added noise can reduce the inversion attack accuracy but at the same time degrades the model training performance \cite{10.5555/3454287.3455674, Ziller2021}. We evaluate the attack and model performances under various differential privacy budgets (noise magnitudes). A typical benchmark NN attack, Deep Leakage Gradients (DLG), is also implemented for this comparison from \cite{zhu2019deepleakagegradients}. 

\subsection{Assumptions}
Assumptions on the setup and the success criteria of the attack are provided below:

\subsubsection{Unknown input data}
In an attack scenario, there is no access to the original underlying data $\mathbf{x}$ neither to the target values $\mathbf{y}$. The initialized weights for the VQNN model gradients are all unit vectors. For robustness, attacks are performed also successfully with random initialized weights (see also Appendix, Fig.\ref{fig:random_initial_weights}). 

\subsubsection{Inversion evaluation}
Inversion success is measured by using Eq. \ref{eq:grad_cost_fun}, the Euclidean distance between the guessed $\mathbf{x}'$ and the real $\mathbf{x}$, $MSE(\mathbf{x}', \mathbf{x})$. There are other metrics to evaluate attacks \cite{avd}, but they are concerned with different types of datasets and embeddings. To evaluate if the algorithm successfully recovers the data, it is sufficient to use a finite and small number of iterations (up to 200) are used in a single-shot trials, and a sample size of 10, to obtain the success rate of the trials. 

\subsubsection{Convergent criteria and a successful attack}
The attack is considered successful if it meets the threshold $MSE(\mathbf{x}',\mathbf{x})\leq 0.005$. To support that this threshold is sufficient to recover the vector $\mathbf{x}$ (at the global minima) with high precision. A convergence analysis of longer runs is also provided in Appendix. Fig. \ref{fig:attacks_kalman_6_dims} . This supporting data shows recovery of all sample with error down to $MSE(\mathbf{x}',\mathbf{x}) \sim 1\mathrm{e}{-10}$ for initial vectors $\mathbf{x}'$ constrained to $\geq 0.005$ random distances.

\subsubsection{Over-parameterized Models}
The inversion is possible as long as the model (the number of equations can be constructed from $\boldsymbol{\theta}$) is comparable or larger than the number of unknowns ($\mathbf{x}$, $\mathbf{y}$)\cite{qian2021minimalmodelstructureanalysis}. In NN, this is usually the case, as models generally contain a large number of parameters. VQNN models are small, and using batch size larger than one is sometimes sufficient to obtain an `under-parameterized' model. In fact, in a batch size larger than one, it requires to use more complex models in order to remain `over-parameterized'. Note, that increasing batch sizes as a strategy to prevent the attack is not sufficient, as it can cause also to large performance degradation. Specifically, in the model used here (Simple Entangled VQNN), increasing the batch size beyond 3 inputs vectors already affects performance dramatically.

\section{Results and Analysis}
\label{sec:results}

\begin{figure}[!t]
    \centering
\includegraphics[width=0.45\textwidth]{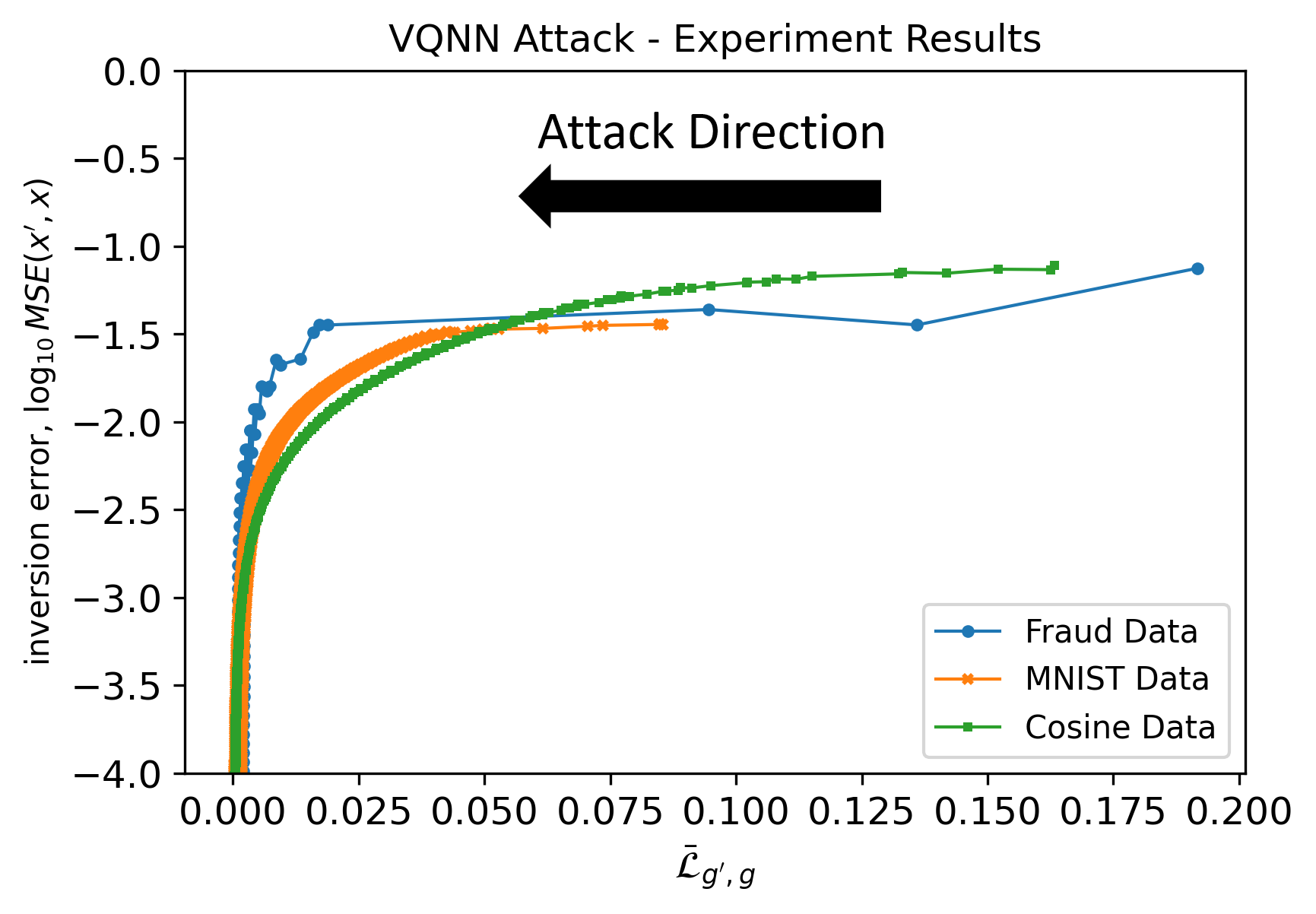} 
\caption{6-dimensional Complex VQNN model (Fig. \ref{fig:architec-models} and $ZZEfficient$) on a variety of datasets. Each line provides a visual of the optimization path (attacks) taken to move the random starting point $\mathbf{x}'$ (start of the line on the right) closer to the baseline/true value $\mathbf{x}$. The $y$-axis is the log MSE distance between the $\mathbf{x}'_{k}$ and the baseline $\mathbf{x}$, with $K$ the number of iterations for the attack. The $x$-axis is the loss function between gradient $\mathbf{g}'$ and original gradient $\mathbf{g}$ (Eq. \ref{eq:grad_cost_fun}).}
\label{fig:overview_attacks}
\end{figure}

\subsection{Attack Results}

Fig.~\ref{fig:overview_attacks} shows successful convergence profiles of the inversion attacks on various datasets for Complex and $ZZEfficient$ models. Averaged error statistics are in Table \ref{tab:attacks}. Using the scheme from Alg. \ref{alg:batch} (without the Kalman component), the adaptive low-pass filter attack can recover all sensitive input data $\mathbf{x}$ in small number of iterations, often within a single trial (except 6 qubits at $40\%$). Results in Table \ref{tab:attacks} reports that adding Kalman improves further the success and convergence rates (see also Appendix Figs. \ref{fig:cosine}, \ref{fig:mnist},  \ref{fig:fraud}). Increasing the number of iterations would reduce error to $1\mathrm{e}{-10}$ (see convergence analysis in Appendix Fig. \ref{fig:attacks_kalman_6_dims}). In Fig. \ref{fig:iterations}, we assess the attack across different circuit depths, showing model complexity's impact. The attack recovers original data in all cases, though convergence takes longer due to increased ansatz complexity. More gates lead to longer optimization but remain fully recoverable as expected.  

\begin{figure}[!t] 
    \centering 
    \includegraphics[width=0.48\textwidth]{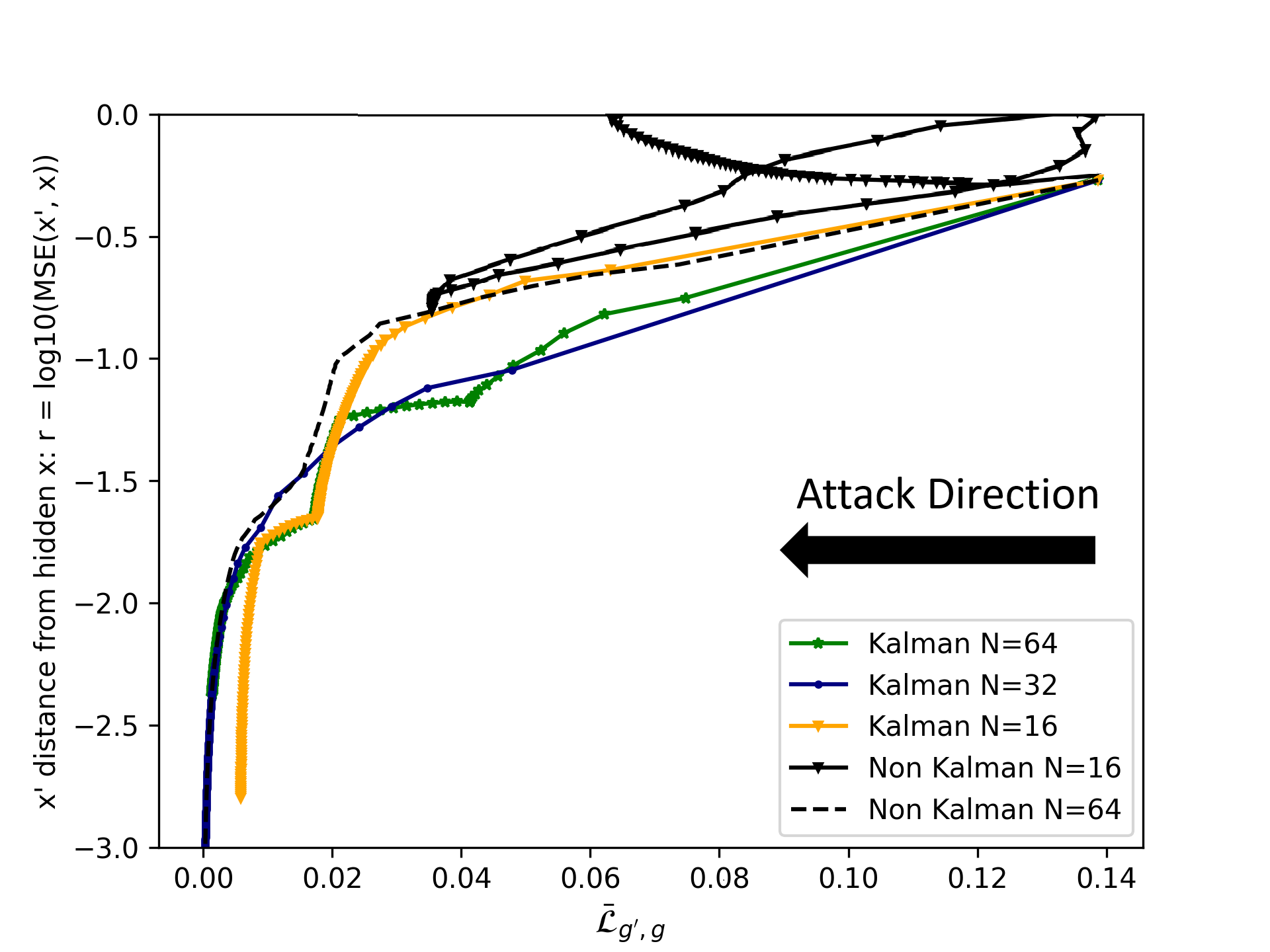} 
    \caption{VQNN complex model $ZZEfficient$ with 2 qubits/inputs $\mathbf{x}$. The graph shows the Kalman based attacks on Fraud data with different number of perturbations $(N=16, 32, 64)$ for the adaptive low-pass filtering (Alg. \ref{alg:batch}). The figure also includes a failed non-Kalman attack for $N=16$ and a successful non-Kalman attack for $N=64$.} 
    \label{fig:kalman_perturbations} 
\end{figure}

In addition, Fig. \ref{fig:kalman_perturbations} shows the increase of efficiency when adding Kalman Filter to update the step in the attack.  With Kalman filter updates we can successfully recover $\mathbf{x}$ with small starting filter sizes, e.g. $N=16$, which improves significantly the convergent rate. It shows that for $N=16$ the non-Kalman attack with adaptive moving average fails to converge at all, but is still successful when using sufficiently large filters, e.g. $N=64$.

\begin{figure}[!t] 
    \centering 
    \includegraphics[width=0.40\textwidth]{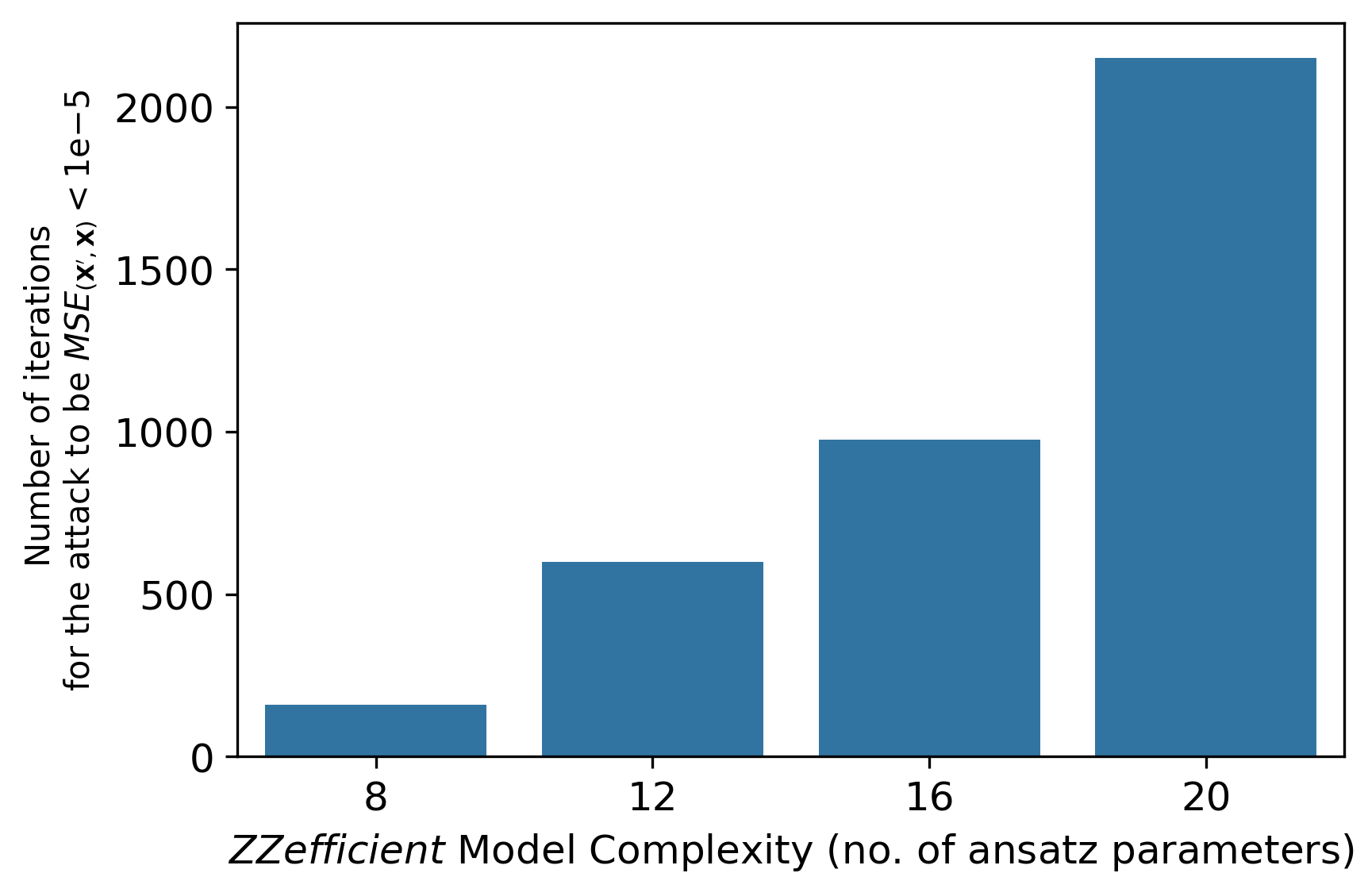}     \caption{The plot shows the number of iterations required for the attack to reach error of $1\mathrm{e}{-5}$ for a $ZZEfficient$ model of 2 dimensions/qubits of different complexities (8, 12, 16, and 20 parameters ansatz). The results show the average iteration number for Cosine and Fraud datasets together.}  
\label{fig:iterations} 
\end{figure}

\subsection{Batch Data}
Since in many ML and FL pipelines the data that the model is trained on a batch of inputs, and subsequently the gradients are a sum of these batches, the model quickly becomes `under-parameterized' in comparison to the size of unknown inputs. Table \ref{tab:attacks} shows that using a batch size of 2, the attack algorithm approximates $\mathbf{x}$ as close as $MSE(\mathbf{x}',\mathbf{x})=2\mathrm{e}{-5}$. When batch size increases to 3 though, the accuracy decreases to $MSE(\mathbf{x}',\mathbf{x})=0.003$. Due to the small size of VQNN in comparison to classical NNs, it is enough to use a batch of size above 3 to not be able to recover the gradients in the VQNN model used. However, as long as the model has sufficient complexity it would possible to recover the data as shown in Table \ref{tab:attacks}. Naturally, increasing the batch size in such small models would also lead to reduced training performance, hence it is not considered as a sufficient defense against this attack. For example, for a Simple Entangled VQNN model trained on the Cosine dataset and batch size 5, the test accuracy is 93\%; with batch size 20 the test accuracy decreases to 88\%; with batch size 60 the accuracy is down to 71\%; and batch size 100 the accuracy is 66\%.

\begin{table*}[!t]
\caption{Train/test model accuracy of a 5-fold Cross Validation set, with the standard errors $(\pm SE)$ of the results. The attacks on the models are performed once. The VQNN models with Fraud data is the $ZZEfficient$ and for the Cosine data is the `Complex entangled' (Fig. \ref{fig:architec-models} ). A typical attack for NN, Deep Leakage Gradients (DLG) \cite{zhu2019deepleakagegradients}, is also included here for NN. }
\label{tab:attacks2}
\resizebox{\textwidth}{!}{
\begin{tabular}{|l|l|l|l|l|}
\hline
 Models & Train accuracy $(\pm SE)$ &Test accuracy $(\pm SE)$ & DLG (mse, iters) & Kalman (mse, iters) \\ \hline
Fraud 6 input 16weights/ NN & 91\% $(\pm 0.8\%)$ & 88\% $(\pm 1.0\%)$ & \textless{}0.001, 10 & \textless{}0.001, 200 \\ 
Fraud 6 inputs 16weights/ NN + Noise0.05 & 90\% $(\pm 0.8\%)$ & 88\% $(\pm 0.9\%)$& \textless{}0.002, 10 & \textless{}0.004, 200 \\ 
Fraud 6 inputs 16weights/ NN + Noise0.14& 80\% $(\pm 1.2\%)$ & 79\% $(\pm 1.5\%)$& \textless{}0.06, 12 & \textless{}0.017, 200 \\ 
Fraud 6 input/ VQNN & 80\% $(\pm 2.0\%)$ & 78\% $(\pm 2.3\%)$ & - & \textless{}0.003, 200 \\ 
CosinBenchmark6 VQNN & 99\% $(\pm 0.2\%)$ & 95\% $(\pm 0.3\%)$ & - & \textless{}0.004, 200 \\ 
CosinBenchmark2 VQNN + Noise0.08& 88\% $(\pm 0.3\%)$ &  86\% $(\pm 0.4\%)$ & -  & \textless{}0.007, 200 \\ 
CosinBenchmark2 VQNN & 99\% $(\pm 0.2\%)$ & 98\% $(\pm 0.5\%)$ & -  & \textless{}0.001, 200 \\ 
CosinBenchmark NN16x2 & 97\% $(\pm 0.2\%)$ & 96\% $(\pm 0.4\%)$& \textless{}0.002, 10 & \textless{}0.001, 200 \\
CosinBenchmark NN16x2 + Noise0.08 & 87\% $(\pm 0.3\%)$ & 85\% $(\pm 0.4\%)$& \textless{}0.01, 10 & \textless{}0.003, 200 \\ \hline
\end{tabular}
}
\end{table*}

\subsection{Comparison to NN and Differential Privacy}

Table \ref{tab:attacks2} summarizes model performance and attack error for NN and VQNN on Fraud and Cosine datasets. The DLG attack is successful for NN but fails for VQNN, hence the results are compared to our algorithm which can successfully invert the data. We also examine the model performance and the attack performance when noise is added to increase privacy and potentially prevent the attack.

\begin{figure}[!t] 
    \centering 
    \includegraphics[width=0.46\textwidth]{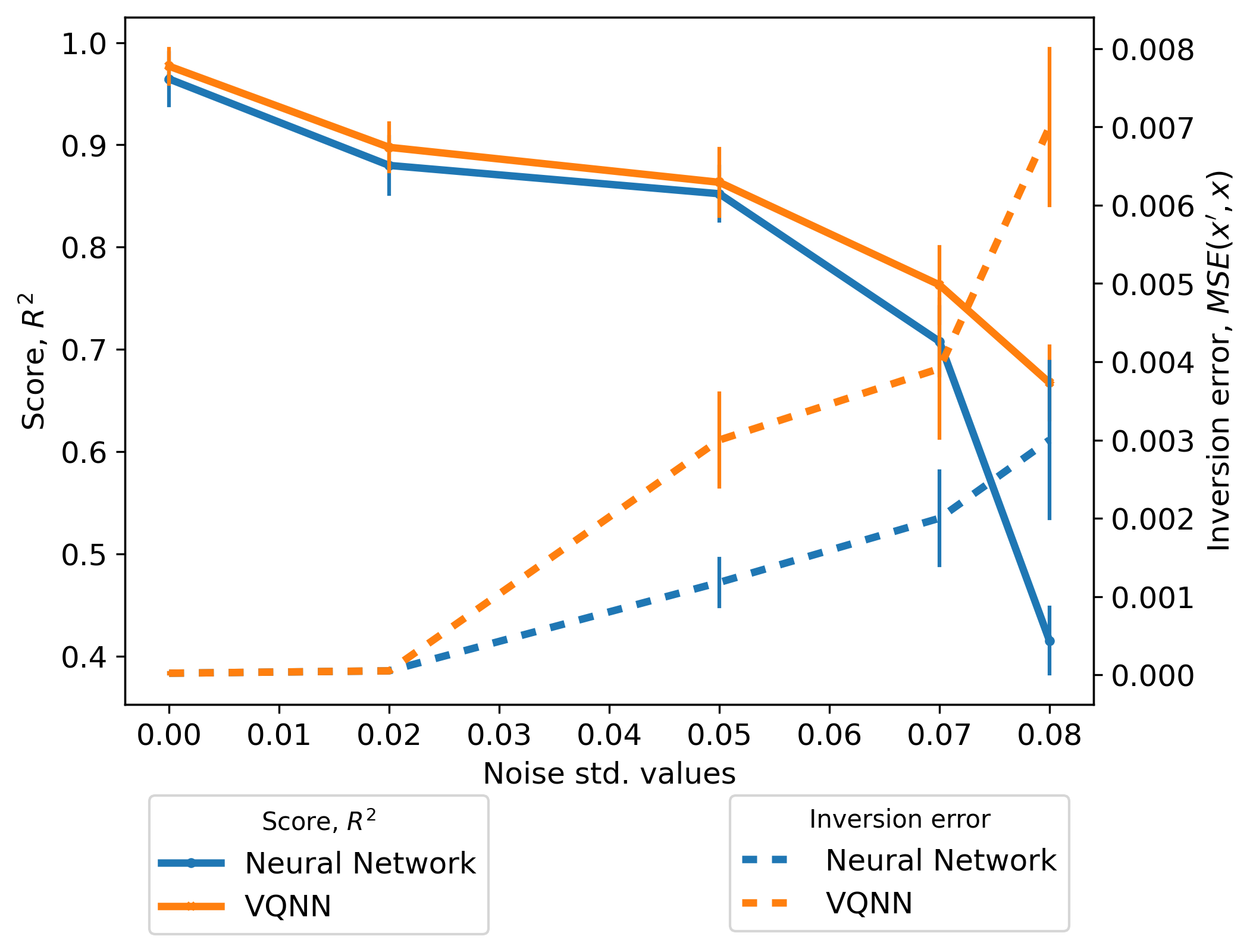} 
    \caption{The plot shows the $R^2$ regression score (LH $y$-axis and solid lines) and the inversion error between $\mathbf{x}'$ and $\mathbf{x}$ (RH $y$-axis and dashed lines) over different noise standard deviation ($\sigma$) values ($\sigma=0.0, 0.02, 0.05, 0.07, 0.08$) added to gradients shared. It follows the Cosine data results from Table \ref{tab:attacks2}. The models are the Cosine Benchmark VQNN 2 qubits and the Cosine Benchmark NN 2 dimensions.}  
\label{fig:training-noise-levels-cosine} 
\end{figure}

\begin{figure}[!t] 
    \centering 
    \includegraphics[width=0.46\textwidth]{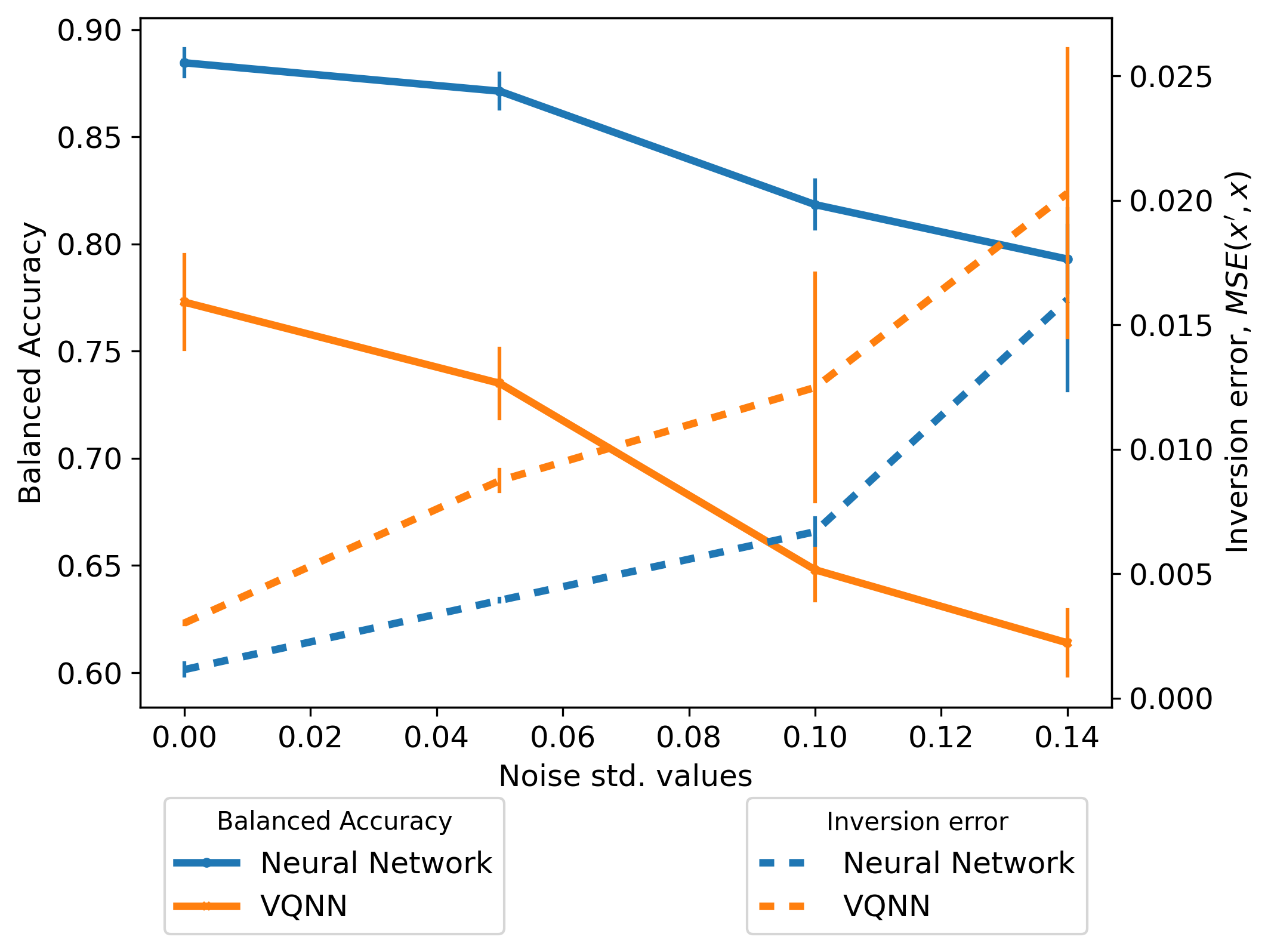} 
    \caption{The plot shows the balanced accuracy (LH $y$-axis and solid lines) and the inversion error between $\mathbf{x}'$ and $\mathbf{x}$  (RH $y$-axis and dashed lines) over the different noise levels ($\sigma= 0.0, 0.05, 0.10, 0.14$) added to the gradients shared. It follows the Fraud data results from Table \ref{tab:attacks2}. The model is the 6 dimensional $ZZEfficient$.}  
\label{fig:training-noise-levels-fraud} 
\end{figure}

Privacy versus performance results are also provided for Cosine (Fig. \ref{fig:training-noise-levels-cosine}) and Fraud datasets (Fig. \ref{fig:training-noise-levels-fraud}). VQNNs excel in fitting cosine functions due to their quantum gate design, outperforming NNs with few linear layers. Fig. \ref{fig:training-noise-levels-cosine} shows VQNNs offer better security against inversion with noise, while NNs remain vulnerable even with degraded performance. The inversion error's square root aligns with added noise, as shown in \cite{Eloul_2024_WACV}.
However for more realistic datasets, such as the Fraud dataset, VQNNs struggle to match simple NNs with 16 weights. Fig. \ref{fig:training-noise-levels-fraud} indicates VQNNs may be more secure but with lower performance. Notably, noise with standard deviation $\sigma=0.1$ yields significant inversion error ($MSE(\mathbf{x}',\mathbf{x})\sim\pm0.013$) while maintaining model utility ($R^2=0.65$) for VQNN. NNs perform better than VQNN but with smaller noise inversion error. With that, other considerations may included in practice. For example, it is worthwhile to note that noise is inherent in quantum hardware, often reduced in VQNN gradients \cite{Kreplin2024reductionoffinite}, but can be leveraged here for privacy. This suggests that advanced low-error quantum computers may not be essential, as noisy systems could benefit privacy. 

\section{Conclusion}
This paper introduces a numerical gradient attack scheme that effectively navigates the frequency modulated landscape of VQNN to obtain global minima via adaptive low-pass filtering. The algorithm proposed can be further optimized with Kalman filter updates to obtain efficient convergence. We show empirically, that as long as the model is trainable (numerically), the proposed attack can invert successfully gradients to data within the precision of the numerical scheme, error down to $\sim 1\mathrm{e}{-10}$. Analysis of various models and datasets show that the algorithm can be used as a benchmark scheme to evaluate the tradeoff between privacy and performance in distributed learning setup, and the practicality of VQNN in comparison to classic NN with differential privacy.
Finally, while this work explores optimization for the inversion of gradients, this optimization approach with low-pass filtering could be insightful for future studies in improving training and the usability of VQNN models.

\section*{Disclaimer}

This paper was prepared for informational purposes by the Global Technology Applied Research center of JPMorgan Chase \& Co. This paper is not a merchandisable/sellable product of the Research Department of JPMorgan Chase \& Co. or its affiliates. Neither JPMorgan Chase \& Co. nor any of its affiliates makes any
explicit or implied representation or warranty and none of them accept any liability in connection with this paper, including, without limitation, with respect to the completeness, accuracy, or reliability of the information contained herein and the potential legal, compliance, tax, or accounting effects thereof. This
document is not intended as investment research or investment advice, or as a recommendation, offer, or solicitation for the purchase or sale of any security, financial instrument, financial product or service, or to be used in any way for evaluating the merits of participating in any transaction.

\ifCLASSOPTIONcaptionsoff
  \newpage
\fi



\bibliographystyle{IEEEtran}
\bibliography{bare_jrnl}

\appendices

\section{Kalman Filter}

The Kalman model's matrices and vectors, as specified in the Kalman filter methods \cite{10.5555/897831, rhudy2017kalman}, are initialized ($k=0$) as all-ones elements ($\mathbf{F}, \mathbf{Q}, \mathbf{P},\mathbf{H}, \mathbf{R}$) and zero initial state. Therefore, we do not assume a prior knowledge of the state or the transitions $\mathbf{F}$. The Kalman filter at each iteration step is divided into two parts, the prediction and the measurement update. The process starts by predicting the state from previous iteration $\mathbf{x}'^{k-1}$ (Eq.\ref{sgd}) using the transition matrix $\mathbf{F}$: 

\begin{align}\label{eq:kalman_predict}
\hat{\mathbf{x}}'^{k-1} &= \mathbf{F} \mathbf{x}'^{k-1}, \\
\hat{\mathbf{P}}^{k-1} &= \mathbf{F} \mathbf{P}^{k-1} \mathbf{F}^T + \mathbf{Q}, \nonumber
\end{align}

where $\mathbf{P}^{k-1}$ is the prior covariance matrix of the state and $\mathbf{Q}$ is the process noise covariance. The update step refines the state estimate using the current measurement $\mathbf{x}'^{k}$ and the predicted previous state $\hat{\mathbf{x}}'^{k-1}$:

\begin{align}\label{eq:kalman_update}
\tilde{\mathbf{x}}'^k &= \mathbf{x}'^k - \mathbf{H} \hat{\mathbf{x}}'^{k-1}, \\ \nonumber
\mathbf{S}^k &= \mathbf{H} \hat{\mathbf{P}}^{k-1} \mathbf{H}^T + \mathbf{R}, \\ \nonumber
\mathbf{D}^k &= \hat{\mathbf{P}}^{k-1} \mathbf{H}^T \mathbf{S}^{-1, k}, \\ \nonumber
\mathbf{x}'^{k} &= \hat{\mathbf{x}}'^{k-1} + \mathbf{D}^k \tilde{\mathbf{x}}'^k, \\ \nonumber
\mathbf{P}^{k} &= (\mathbf{I} - \mathbf{D}^k \mathbf{H})\hat{\mathbf{P}}^{k-1}, \\ \nonumber
\end{align}

where $\mathbf{H}$ is the measurement matrix, $\mathbf{R}$ is the measurement noise covariance, $\mathbf{S}^k$ the innovation covariance, and $\mathbf{D}^k$ is the Kalman gain. 
 
\section{Supporting Analysis}

\begin{figure}[!t]
    \centering
\includegraphics[width=0.45\textwidth]{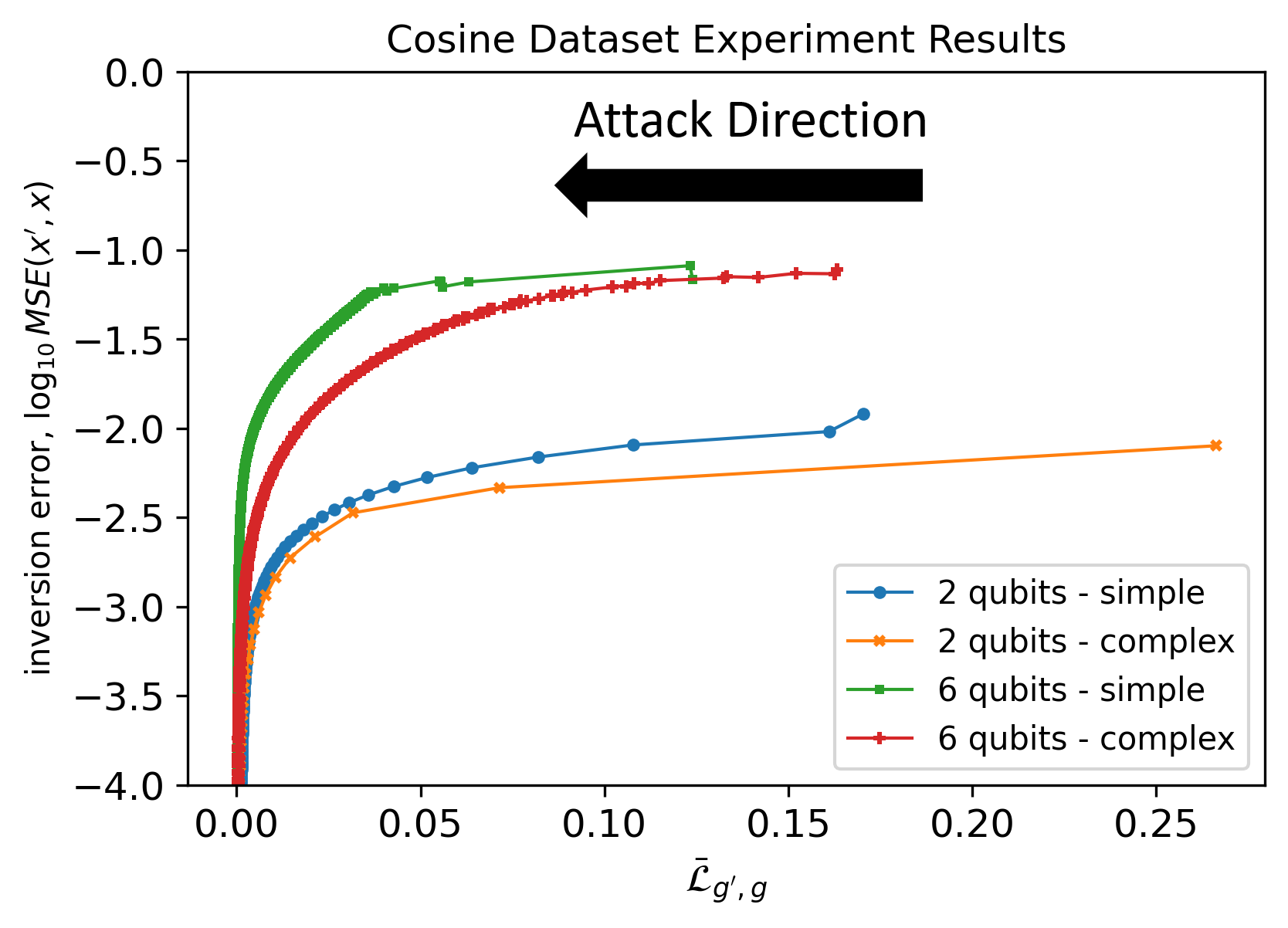} 
\caption{These plots feature 2 and 6-dimensional Cosine datasets $\mathbf{x}$. Fig. \ref{fig:architec-models} shows the Simple and Complex models respectively. Each line provides a visual of the optimization path taken to move the random point $\mathbf{x}'$ (start of the line on the right) closer to the original $\mathbf{x}$ (at coordinates $x$-axis$=0$, $y$-axis$=0$). The $y$-axis is the $log10$ MSE distance between the $\mathbf{x}'_{k}$ and the $\mathbf{x}$, with $k=\{1,2,\cdots, K\}$ the attack steps and $K$ the total steps. The $x$-axis is the distance between the gradient $\mathbf{g}_{k}'$ of the $\mathbf{x}'_{k}$ and the gradient $\mathbf{g}$}
\label{fig:cosine}
\end{figure}

Fig.\ref{fig:cosine} shows the attacks on cosine generated dataset, for 2 and 6 dimensions. The models are the Simple and the Complex from Fig. \ref{fig:architec-models}. The attack gets as close as $MSE(\mathbf{x}', \mathbf{x})$ is $\leq 1\mathrm{e}{-10}$.

\begin{figure}[!t]
    \centering
\includegraphics[width=0.45\textwidth]{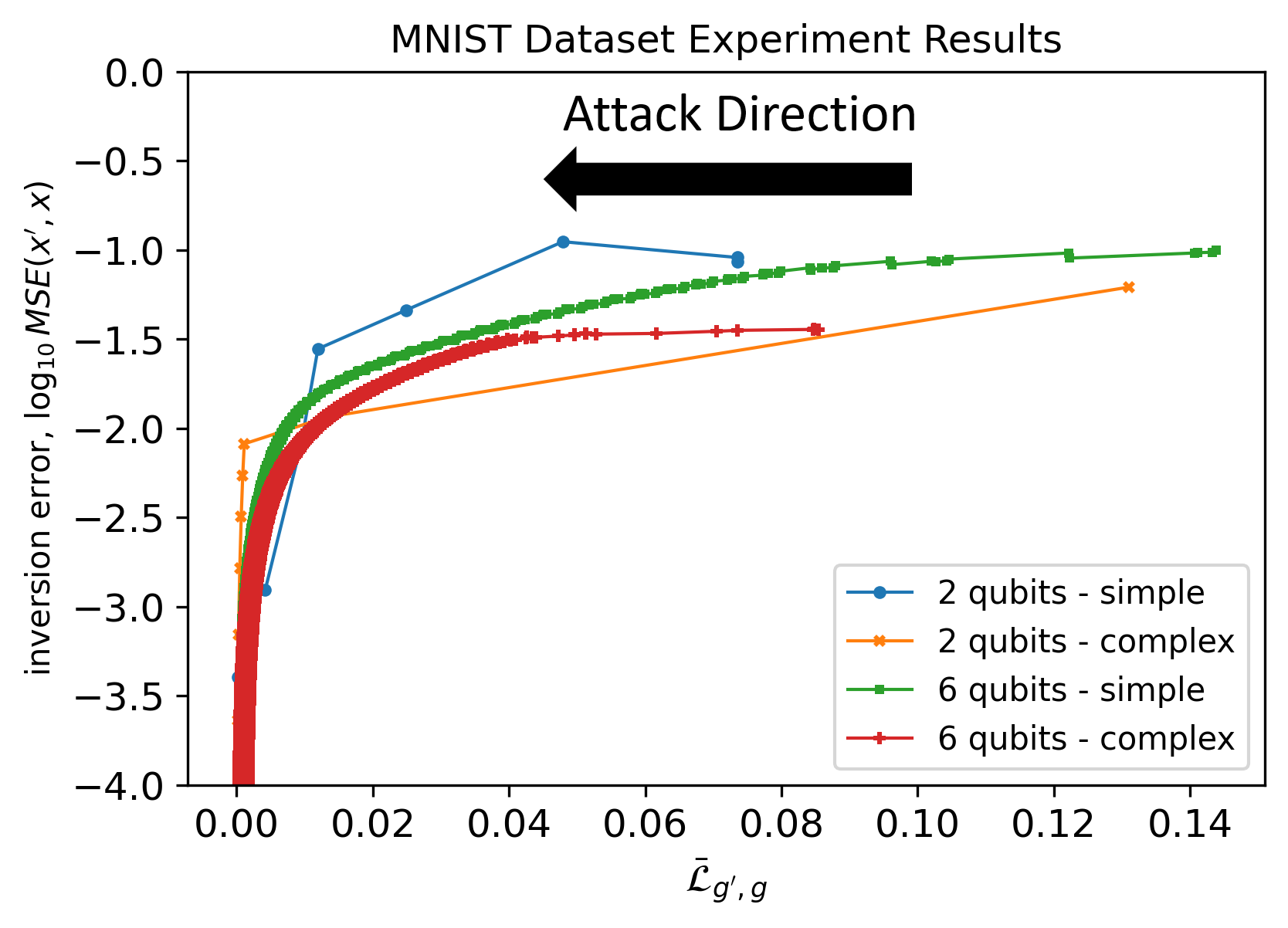} 
\caption{These plots feature 2 and 6-dimensional MNIST data. Fig. \ref{fig:architec-models} shows the Simple and Complex models respectively. Each line provides a visual of the optimization path taken to move the random point $\mathbf{x}'$ (start of the line on the right) closer to the original $\mathbf{x}$ (at coordinates $x$-axis$=0$, $y$-axis$=0$). The $y$-axis is the $log10$ MSE distance between the $\mathbf{x}'_{k}$ and the $\mathbf{x}$, with $k=\{1,2,\cdots, K\}$ the attack steps and $K$ the total steps. The $x$-axis is the distance between the gradient $\mathbf{g}_{k}'$ of the $\mathbf{x}'_{k}$ and the gradient $\mathbf{g}$}
\label{fig:mnist}
\end{figure}

Fig.\ref{fig:mnist} pictures the attack on the MNIST dataset \cite{Lecun726791}. The $MSE(\mathbf{x}', \mathbf{x})$ error of the attacks is $\leq 1\mathrm{e}{-10}$. The models are the Simple and the Complex from Fig. \ref{fig:architec-models}. 

\begin{figure}[!t]
    \centering
\includegraphics[width=0.45\textwidth]{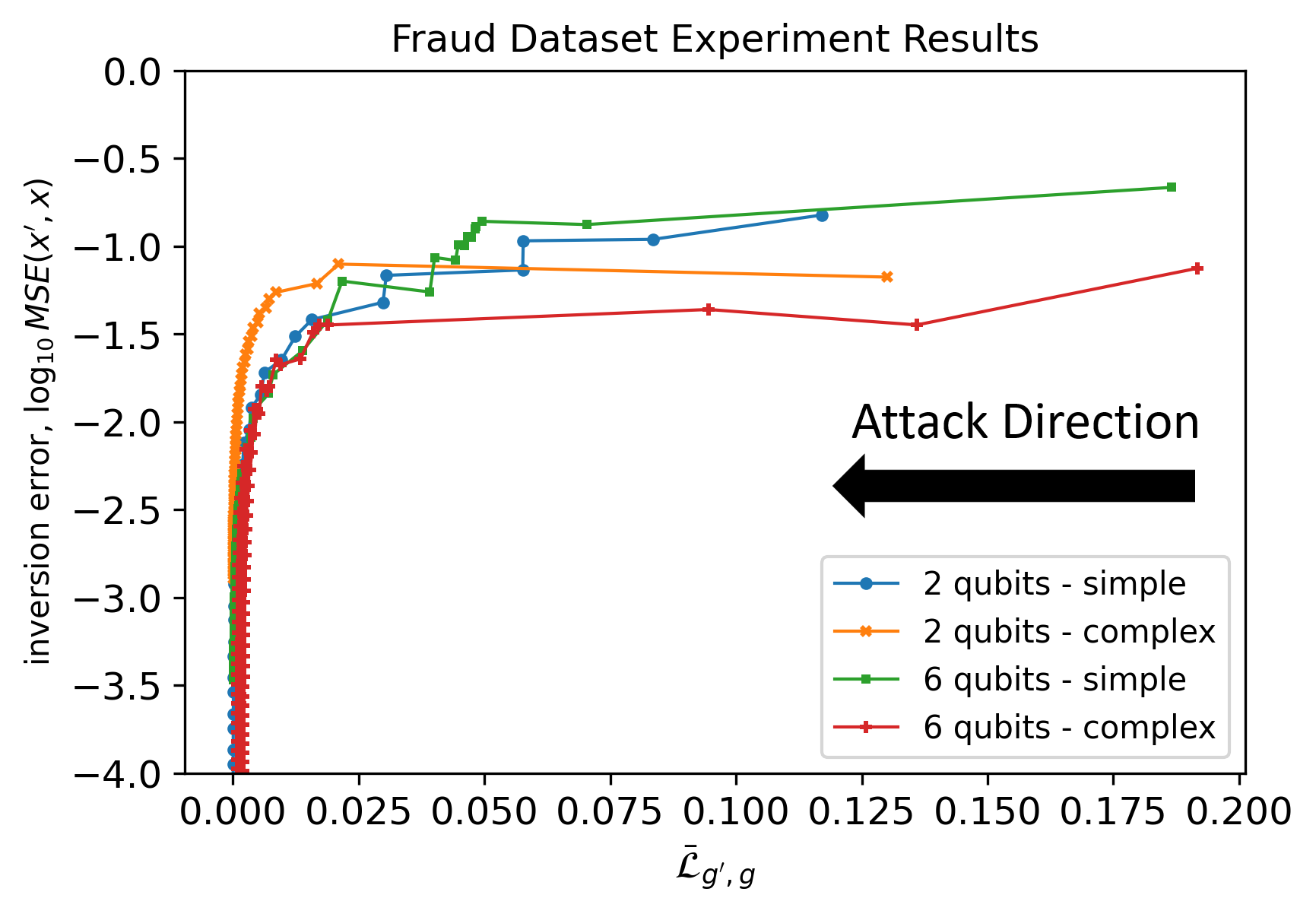} 
\caption{These plots feature 2 and 6-dimensional credit Fraud classification data \cite{ULB_2018}. \textbf{For the complex model we relied on the trainable architecture $ZZFeatureMap$ \cite{arxiv.2207.11449, arxiv.2408.10274} and $EfficientSU2$ \cite{arxiv.2402.16465}}. Each line provides a visual of the optimization path taken to move the random point $\mathbf{x}'$ (start of the line on the right) closer to the original $\mathbf{x}$ (at coordinates $x$-axis$=0$, $y$-axis$=0$). The $y$-axis is the $log10$ MSE distance between the $\mathbf{x}'_{k}$ and the $\mathbf{x}$, with $k=\{1,2,\cdots, K\}$ the attack steps and $K$ the total steps. The $x$-axis is the distance between the gradient $\mathbf{g}_{k}'$ of the $\mathbf{x}'_{k}$ and the gradient $\mathbf{g}$}
\label{fig:fraud}
\end{figure}

Fig. \ref{fig:fraud} illustrates the attacks on the Fraud dataset \cite{ULB_2018}. The VQNN models are the Simple and for complex the $ZZEfficient$ (Fig. \ref{fig:architec-models}). The $MSE(\mathbf{x}', \mathbf{x})$ is $\leq 1\mathrm{e}{-10}$. 

\begin{figure}[!t]
    \centering
    \includegraphics[width=0.42\textwidth]{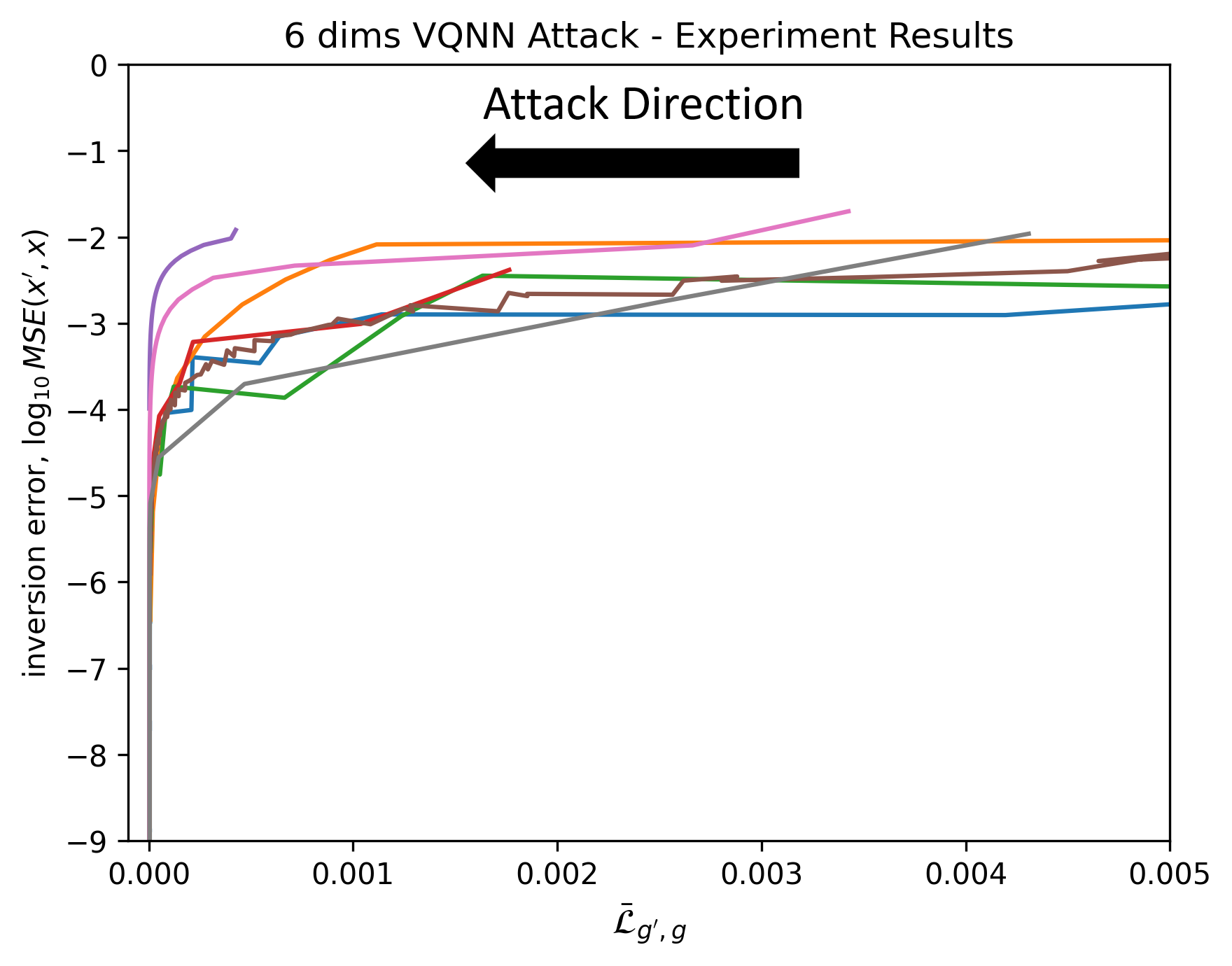} 
    \caption{Attack on Fraud dataset of the 6 dimensional VQNN $ZZEfficient$ model. We show that when we attack the 6-VQNN model for more than $k\geq 200$ the results will converge to as low as $1\mathrm{e}{-10}$.}
    \label{fig:attacks_kalman_6_dims}
\end{figure}

Fig. \ref{fig:attacks_kalman_6_dims} shows that for runs with more that $k>200$ iterations the algorithm eventually converges to as low as $MSE(\mathbf{x}',\mathbf{x})=1\mathrm{e}{-10}$.

\begin{figure}[!t]
    \centering
    \includegraphics[width=0.42\textwidth]{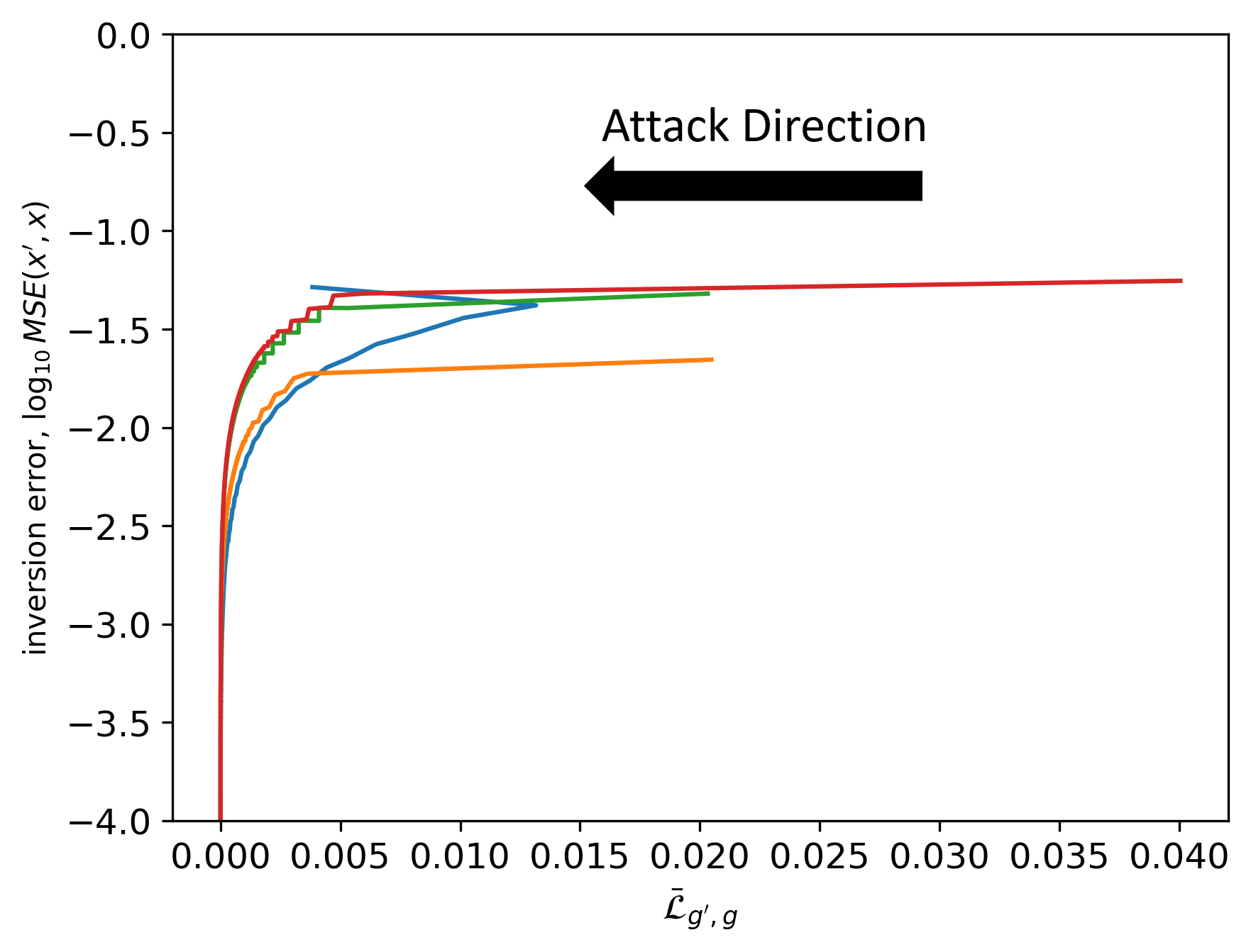} 
    \caption{We perform 4 attacks by using different initial model weights each time. We utilize a 2-qubit/inputs Simple model (\ref{fig:architec-models}) on Cosine data. The results converge to as low as $1\mathrm{e}{-10}$.}
    \label{fig:random_initial_weights}
\end{figure}

Fig. \ref{fig:random_initial_weights} depicts attacks on the Cosine data by using different initialized weights for each attack attempt. It proves the robustness of the attack algorithm under different scenarios (instead of starting the weights with 1s).

\begin{figure}[!t]
    \centering
    \includegraphics[width=0.42\textwidth]{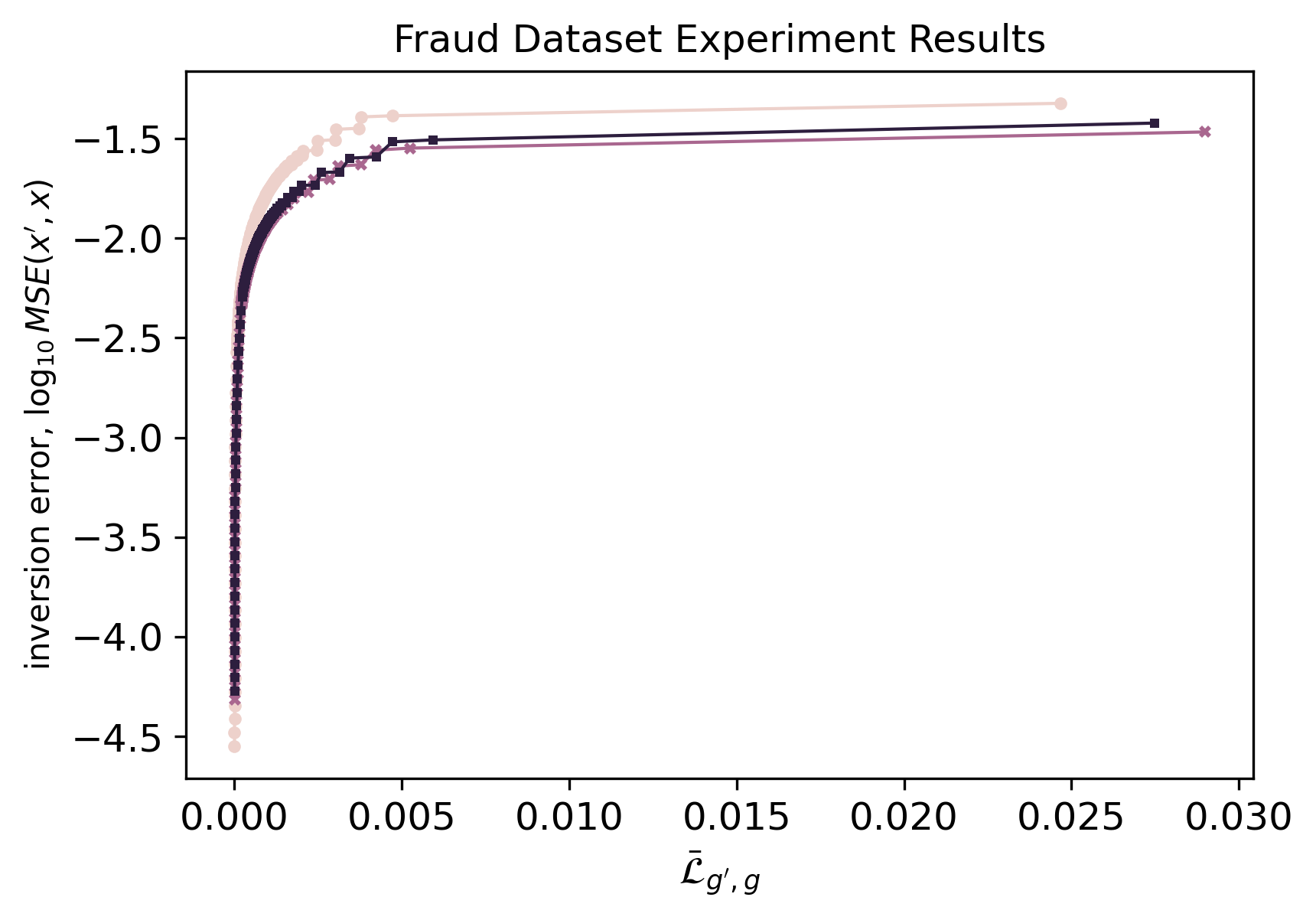} 
    \caption{We perform 3 attacks using as a base gradient $\mathbf{g}$ the gradient after the successful training of a VQNN classification model (Fraud data, $ZZEfficient$, 6 qubits/dimensions). The results converge to as low as $1\mathrm{e}{-5}$.}
    \label{fig:trained_model_attack}
\end{figure}

In Fig. \ref{fig:trained_model_attack}, we observe that the attack algorithm is successful even when we target gradients coming from a trained model $\mathbf{g}$. This shows that in real-life FL settings the attack algorithm can approximate the base data $\mathbf{x}$ under most scenarios we have put forward.  

\clearpage
\section{List of Symbols}

The following list contains the symbols used in the text:

\begin{itemize}
    \item $\boldsymbol{\theta}$: Vector of ansatz parameters/weights of the Quantum Machine Learning model. 
    \item $M$: The total number of weights of a QVNN model.
    \item $C$: The total number of classes in the target values $\mathbf{y}$.
    \item $ \mathrm{Loss}$: The loss function of the Quantum Machine Learning model.
    \item $\mathbf{g}$: Vector of gradient $\partial  \mathrm{Loss}/\partial\boldsymbol{\theta}$.
    \item $\mathcal{L}_{\mathbf{g}', \mathbf{g}}$: The loss function of the Quantum Machine Learning model gradient $\mathbf{g}$ and an approximate gradient $\mathbf{g}$.
    \item $\nabla$: The symbol of gradient. 
     \item $\mathbf{x}$: A vector of the input data that feed into the Quantum Machine Learning model. 
     \item $\mathbf{y}$: The target values/classes that are used during the training process.
     \item $\hat{\mathbf{y}}$: Predicted values.
     \item $I$: The total number of training steps/epochs. The total number of the data points.
     \item $i$: The training steps (iterations) during the training processes $i \in \{1,2,\dots,I\}$.
     \item $B$: The batch data size. 
     \item $b$: The number of the batch segment.
     \item $\gamma$: The learning rate parameter during model training/optimization. 
     \item $C_l$: The total number of clients in a Federated Learning framework. 
     \item $(')$: We use the apostrophe/quote symbol to designate proxy elements, ($\mathbf{x}'$ proxy to original data $\mathbf{x}$ or $f'$ proxy to original model $f$).
     \item $J$: The number of features/dimensions. 
     \item $f$: A function that takes inputs $\mathbf{x}$ and parameters $\boldsymbol{\theta}$ and generates predicted values $\hat{\mathbf{y}}$.
     \item $\hat{\mathbf{e}}$: A unit vector.
     \item $\delta$: A parameter for the finite difference method that approximates gradients.
     \item $\eta$: The learning rate of the gradient descent algorithm.
     \item $q$: The number of qubits of a VQNN model. For this paper, all of our cases are $J=q$. Number of qubits equals the number of inputs/dimensions of $\mathbf{x}$.
     \item $\Phi$: The feature map of a VQNN model .
     \item $U_\Phi$: The unitary operator of a VQNN model.
     \item $\psi$: The VQNN function that gives us the measurement using the parameterized circuit.
     \item $V$: The ansatz of a VQNN model.
     \item $\hat{O}$: A Hermitian operator used as an observable in the VQNN.
     \item $\otimes$: Tensor product.
\end{itemize}

\end{sloppypar}

\end{document}